\documentclass{article}

\usepackage[preprint]{neurips_2025}

\usepackage[utf8]{inputenc} 
\usepackage[T1]{fontenc}    
\usepackage{hyperref}       
\usepackage{url}            
\usepackage{booktabs}       
\usepackage{amsfonts}       
\usepackage{nicefrac}       
\usepackage{microtype}      
\usepackage{xcolor}         

\usepackage{amsmath}
\usepackage{amssymb}
\usepackage{multirow}
\usepackage{amssymb}   
\usepackage[dvipsnames, svgnames, x11names]{xcolor}
\usepackage{booktabs}       
\usepackage{amsfonts}       
\usepackage{nicefrac}       
\usepackage{microtype}      
\usepackage{xcolor}        
\usepackage{wrapfig}    
\definecolor{gray}{rgb}{0.5,0.5,0.5}
\definecolor{cmblu}{RGB}{51,102,240}
\definecolor{cmred}{RGB}{241,22,22}
\usepackage[table]{xcolor} 
\usepackage{colortbl}      
\usepackage{graphicx} 
\usepackage{caption}  
\usepackage{color}
\usepackage{xspace}
\usepackage{amssymb}   
\usepackage{bbding}
\newcommand{\best}[1]{\cellcolor[rgb]{0.957,0.71,0.706}#1}
\newcommand{\besttwo}[1]{\cellcolor[rgb]{0.976,0.859,0.718}#1}
\newcommand{\bestthree}[1]{\cellcolor[rgb]{1,1,0.613}#1}

\title{Diffusion-Driven Progressive Target Manipulation for Source-Free Domain Adaptation}

\author{%
Yuyang Huang\textsuperscript{1}\footnotemark[1], \quad 
Yabo Chen\textsuperscript{1}\footnotemark[1], \quad  
Junyu Zhou\textsuperscript{1}, \quad 
Wenrui Dai\textsuperscript{1}\footnotemark[2], \quad 
Xiaopeng Zhang\textsuperscript{2}\footnotemark[2], \quad  \\
\textbf{Junni Zou}\textsuperscript{1}\footnotemark[2] \quad, 
\textbf{Hongkai Xiong}\textsuperscript{1}, \quad   \textbf{Qi Tian}\textsuperscript{2} \\ 
\textsuperscript{1}Shanghai Jiao Tong University, Shanghai, China \quad
\textsuperscript{2}Huawei Inc., Shenzhen, China\\
\texttt{\{huangyuyang, chenyabo, blabla, daiwenrui, zoujunni, xionghongkai\}@sjtu.edu.cn}\\
\texttt{zxphistory@gmail.com}, \texttt{tian.qi1@huawei.com}\\
\footnotesize *These authors contributed equally to this work.
\footnotesize Corresponding authors: Wenrui Dai; Xiaopeng Zhang; Junni Zou.
}

\begin{document}
\maketitle
\begin{abstract}
Source-free domain adaptation (SFDA) is a challenging task that tackles domain shifts using only a pre-trained source model and unlabeled target data. Existing SFDA methods are restricted by the fundamental limitation of source-target domain discrepancy. Non-generation SFDA methods suffer from unreliable pseudo-labels in challenging scenarios with large domain discrepancies, while generation-based SFDA methods are evidently degraded due to enlarged domain discrepancies in creating pseudo-source data. To address this limitation, we propose a novel generation-based framework named Diffusion-Driven Progressive Target Manipulation (DPTM) that leverages unlabeled target data as references to reliably generate and progressively refine a pseudo-target domain for SFDA. Specifically, we divide the target samples into a trust set and a non-trust set based on the reliability of pseudo-labels to sufficiently and reliably exploit their information. For samples from the non-trust set, we develop a manipulation strategy to semantically transform them into the newly assigned categories, while simultaneously maintaining them in the target distribution via a latent diffusion model. Furthermore, we design a progressive refinement mechanism that progressively reduces the domain discrepancy between the pseudo-target domain and the real target domain via iterative refinement. Experimental results demonstrate that DPTM outperforms existing methods by a large margin and achieves state-of-the-art performance on four prevailing SFDA benchmark datasets with different scales. Remarkably, DPTM can significantly enhance the performance by up to 18.6\% in scenarios with large source-target gaps.
\end{abstract}

\section{Introduction}

Deep learning has achieved remarkable success under the independent and identically distributed (i.i.d.) assumption. However, it suffers from significantly degraded performance on out-of-distribution (OOD) data due to domain shifts. Unsupervised domain adaptation (UDA) mitigates this issue by aligning feature distributions between labeled source and unlabeled target domains, but has to access both datasets during adaptation~\cite{wang2018deep,wilson2020survey}. Source-free domain adaptation (SFDA) considers a more practical but challenging scenario where only the pre-trained source model and unlabeled target data are available~\cite{li2024comprehensive,fang2024source}, and precludes access to source samples during adaptation.

Existing SFDA methods can be primarily classified into non-generation and generation-based methods, with both exhibiting inherent limitations on practical effectiveness. Non-generation methods~\cite{plue,crs,cpd,tpds,difo,prode} predominantly rely on pseudo-labels generated by the source model, and categorize them into a small subset of reliable pseudo-labels and a predominant subset of unreliable pseudo-labels based on certainty metrics~\cite{liang2021source,wang2021source,ding2023proxymix,yang2022divide,yang2023casting,chu2023adversarial,zhang2022divide}. The unreliable pseudo-labels contain substantial label noise, and cannot be easily exploited to extract useful information or infer correct labels through refinement processes. Unfortunately, the amount of label noise is inherently determined by the degree of domain shift between source and target domains~\cite{elr,litrico2023guiding}. This fundamental limitation severely compromises the effectiveness of non-generation methods in challenging scenarios with large domain gaps, and results in significantly unstable performance across different adaptation tasks. For instance, empirical results~\cite{prode} show that, when deploying the same source model across different target domains, significant performance discrepancies emerge (\emph{e.g.}, accuracy over 90\% for Ar$\to$Cl vs. about 60\% for Ar$\to$Pr in the Office-Home dataset). These results underscore the critical sensitivity of non-generation methods to domain shift. 

Generation-based SFDA methods primarily operate at the data level~\cite{cpga,asoge,improvedsfda,ps}. Although these methods could theoretically circumvent the limitation of non-generation methods by avoiding directly using unreliable pseudo-labels, most of them still fail to escape from this domain shift limitation due to the problematic paradigm that generates a pseudo-source domain to convert the SFDA task into a conventional UDA task. The restriction caused by the domain shift between the pseudo-source and target domains is not addressed. Moreover, the generation process often incorporates irrelevant domain features that could further enlarge the discrepancy between the source and target domains for the pseudo-source domain. Consequently, these methods suffer from unsatisfactory performance due to domain shifts.

In this paper, we reveal that existing SFDA methods are fundamentally limited by source-target domain shifts. To break this bottleneck, we resort to a novel generation-based paradigm that directly generates the pseudo-target domain. We propose a Diffusion-Driven Progressive Target Manipulation (DPTM) framework that reliably generates and progressively refines the pseudo target domain to reduce the domain discrepancy from the real target domain and address the fundamental limitation on domain shift for existing methods. The proposed method is shown to achieve remarkable performance gains in challenging DA scenarios with large source-target domain shifts.

To sufficiently and reliably exploit the pseudo-label information, we partition the target data into a trust set and a non-trust set based on the prediction uncertainty of the target model initialized using the source model. For the trust set with low uncertainty, we directly adopt pseudo-labels as supervisory signals for training the target model, following prior works that have established their reliability~\cite{liang2021source,wang2021source,ding2023proxymix,yang2022divide,yang2023casting,chu2023adversarial,zhang2022divide}. Moreover, we also exploit the rich information about the target distribution contained in the potentially unreliable samples from the non-trust set. We uniformly assign a new category label to each of them to prevent potential class imbalance and employ a manipulation strategy to semantically transform each sample toward its newly assigned label while preserving its target-domain features with a latent diffusion model~\cite{rombach2022high}. The manipulated samples simultaneously turn their assigned labels into useful supervisory signals and keep aligned with the target distribution to enhance the adaptation of target models. Furthermore, we propose a Progressive Refinement Mechanism that iteratively refines the pseudo-target domain as well as the target model to progressively reduce the residual label noise due to imperfect pseudo-labeling and the accumulated domain discrepancy caused by the manipulated non-trust set. This significantly diminishes the quantity of non-trust samples and thereby mitigates overall domain shift.

To be concrete, the proposed manipulation strategy of non-trust samples consists of three components. Firstly, as the sampling starting point of diffusion models has been proven to significantly influence the generated image~\cite{wu2023freeinit,qiu2023freenoise,koo2024flexiedit,xu2025stylessp,everaert2024exploiting}, we propose a Target-guided Initialization Mechanism to construct the starting point for sampling  by simultaneously considering the target domain features of the non-trust sample and isolating its semantic leakage that might disturb the semantic transformation. Secondly, we propose a Semantic Feature Injection Mechanism that iteratively injects semantics related to the assigned label into the latent throughout the sampling trajectory via DDIM inversion~\cite{song2020denoising,bai2024zigzag} to ensure the semantic transformation without introducing unrelated domain features. Finally, for consistency of manipulated samples with the target distribution, we present a Domain-specific Feature Preservation Mechanism to actively inject target domain features with an adaptively perturbed latent drawn from the original non-trust sample. 

Experimental results demonstrate that the proposed method achieves superior performance compared to state-of-the-art (SOTA) methods across four standard SFDA benchmarks of different scales. Remarkably, our method successfully overcomes the limitations of existing methods in challenging domain adaptation scenarios involving large domain shifts. For instance, we achieve a gain of \textbf{9.3\%} on D$\to$A and
\textbf{8.2\%} on W$\to$A tasks over the existing SOTA method on the small-scale Office-31 dataset, and a remarkable gain of \textbf{18.6\%} over SOTA for the Rw$\to$Cl task on the medium-scale Office-Home dataset. On the large-scale DomainNet-126 dataset, we achieve a gain of \textbf{24.4\%} over the existing generation-based method and \textbf{6.3\%} over SOTA for the C$\to$P task.

The contributions of this paper are summarized as follows.
\begin{itemize}
\item We propose DPTM, a novel framework for Source-free Domain Adaptation (SFDA) that progressively constructs and refines a pseudo-target domain by leveraging unlabeled target data as references with a latent diffusion model. 
\item We develop a manipulation strategy that leverages Target-guided Initialization, Semantic Feature Injection, and Domain-specific Feature Preservation to semantically transform the unreliable sample toward the newly assigned label while preserving target-domain features.
\item We design a Progressive Refinement Mechanism that progressively reduces the domain discrepancy between the pseudo target domain and the real target domain via iterative refinement.
\end{itemize}

\section{Related Work} 

\textbf{Source-Free Domain Adaptation.} Source-free domain adaptation (SFDA) methods can be broadly categorized into non-generation and generation-based methods. Non-generation methods~\cite{shot,nrc,gkd,hcl,ada,adacon,cowa,sclm,elr,plue,crs,cpd,tpds,difo,prode} mainly employ self-training techniques using pseudo-labels predicted by the source model. However, the inherent unreliability of pseudo-labels caused by source-target domain shifts substantially limits their performance, especially in challenging DA scenarios with significant domain discrepancies. Generation-based methods~\cite{plue,crs,cpd,tpds,difo,prode} usually generate pseudo-source domains to convert SFDA into a conventional UDA problem, but their performance remains constrained by the domain shift between the generated pseudo-source and target domains. Different from existing methods, we develop a novel method that directly generates pseudo-target domains and progressively reduces the domain shift between pseudo-target and real-target samples through iterative refinement to overcome the performance limitations.

\textbf{Diffusion Models.} Diffusion models~\cite{ho2020denoising,dhariwal2021diffusion,rombach2022high} have become state-of-the-art in many generative tasks~\cite{xie2024sanaefficienthighresolutionimage,zheng2024cogview3,zhang2023controlvideo,zhang2024videoelevator,khachatryan2023text2video,imzero,chen2024cascade}. Their exceptional generation capabilities and pre-trained visual knowledge have been successfully transferred to other vision tasks such as image segmentation~\cite{zhang2023tale,zhao2023unleashing} and domain generalization~\cite{huang2024domainfusion,thomas2025s}. In this work, we leverage diffusion models to facilitate SFDA tasks by semantically transforming unreliable target samples toward their assigned category labels while rigorously preserving their target domain characteristics.

Diffusion models are latent variable generative models defined by a forward and reverse Markov process~\cite{ho2020denoising,dhariwal2021diffusion}. The forward process $\{q_t\}_{t \in[0, T]}$ progressively adds Gaussian noise to the data $x_0 \sim q_0(x_0)$ by $q(x_t|x_0)=\mathcal{N}(x_t ; \alpha_t x_0, \sigma_t^2 \mathbf{I})$, where the scheduling hyper-parameters $\alpha_t^2+\sigma_t^2=1$. 
The reverse process $\{p_t\}_{t \in[0, T]}$ gradually removes noise using a learned denoiser $\epsilon_\theta$. Starting from $p(x_T)=\mathcal{N}(\mathbf{0}, \mathbf{I})$, it reconstructs $x_0$ through transitions $p_\theta(x_{t-1}|x_t)=\mathcal{N}(x_{t-1} ; x_t-\epsilon_\theta(x_t, t), \sigma_t^2 \mathbf{I})$. Conditional generation is achieved by incorporating condition $y$ into the denoising process as an input to $\epsilon_\theta\left(x_t, y, t\right)$. Classifier-free guidance enables conditional generation by combining conditional and unconditional denoising predictions with a guidance scale $\gamma_1$ :
\begin{equation}\label{denoising}
\bar{\epsilon}_\theta\left(x_t, y,t\right)=(1+\gamma_1)\epsilon_\theta\left(x_t, y,t\right)-\gamma_1\epsilon_\theta\left(x_t,\varnothing, t\right). 
\end{equation}

\section{Method}

\begin{figure}[!t]
\centering
    \includegraphics[width=\textwidth]{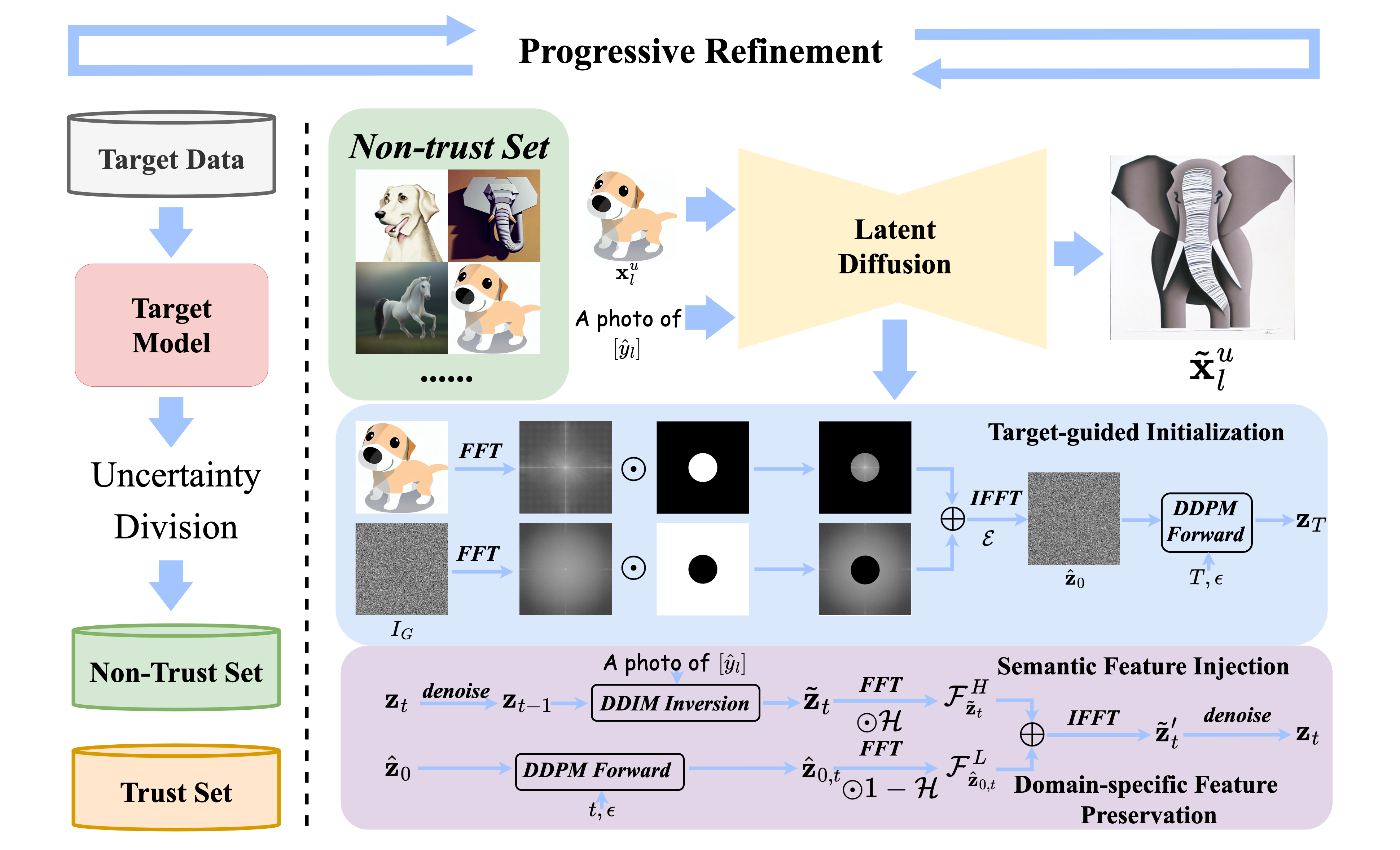} 
    \caption{In DPTM, we employ progressive refinement $R$ times: First, we use the target model to make predictions on the target data. Based on each sample's prediction uncertainty, we divide the target data into a trust set and a non-trust set. For the low-uncertainty trust set, we train the target model using pseudo-labels in a supervised manner. For the high-uncertainty non-trust set, we assign a label $\hat{y}_l$ for each sample $\mathbf{x}_l^u$, employ a manipulation strategy that semantically transforms $\mathbf{x}_l^u$
  toward class $\hat{y}_l$, while preserving the target-domain features of $\mathbf{x}_l^u$. Our manipulation consists of three components: Target-guided Initialization to obtain an effective sampling starting point, Semantic Feature Injection to convert the semantics of the generated sample to $\hat{y}_l$, and Domain-specific Feature Preservation to maintain the generated sample within
the target distribution.}\label{fig:framework} %
\end{figure}

\subsection{Overall Framework}

Let $\mathcal{D}_{src}$ a labeled source domain with input space $\mathcal{X}_{src} = \{\mathbf{x}_i^{src}\}_{i=1}^{N_{src}}$ and label space $\mathcal{Y}_{src} = \{ y_i^{src}\}_{i=1}^{N_{src}}$, and $\mathcal{D}_{trg}$ an unlabeled target domain with input space $\mathcal{X}_{trg} =\{\mathbf{x}_j^{trg}\}_{j=1}^{N_{trg}}$, where $N_{src}$ and $N_{trg}$ denote the number of samples in the source and target domains, respectively. In SFDA, we first train a source model $\phi_{src}: \mathcal{X}_{src} \rightarrow \mathcal{Y}_{src} $ on $\mathcal{D}_{src}$ via supervised learning, and then utilize $\phi_{src}$ and the unlabeled $\mathcal{X}_{trg}$ to learn a target model $\phi_{trg}: \mathcal{X}_{trg} \rightarrow \mathcal{Y}_{trg} $ that generalizes well on $\mathcal{D}_{trg}$.

Figure~\ref{fig:framework} depicts the proposed framework that comprises three key components, including the partition of trust and non-trust sets for unlabeled target data in Section~\ref{targetdivision}, manipulation strategy of the non-trust set in Section~\ref{nottrustmanipulation}, and a progressive refinement mechanism that continuously minimizes the discrepancy between the evolving pseudo-target domain and the real target domain in Section~\ref{updatemechanism}. We initialize the target model $\phi_{trg}$ with the pre-trained source model $\phi_{src}$. The target domain data is first partitioned into a trust set $\mathcal{V}$ and a non-trust set $\mathcal{U}$ based on prediction uncertainty. For any trust sample $\mathbf{x}_k^v \in \mathcal{V}$, we use the corresponding pseudo-label $y_k^p$ as the supervision signal to train $\phi_{trg}$. The non-trust set undergoes diffusion-based manipulation to produce $\mathcal{U}^m$, which is combined with $\mathcal{V}$ to form the pseudo-target domain $\mathcal{D}_p = \mathcal{V} \cup \mathcal{U}^m$. Finally, we optimize the source model $\phi_{src}$ on this pseudo-target domain in a supervised manner, obtaining a target model $\theta_v$.

\subsection{Trust and Non-trust Partition for Target Domain} \label{targetdivision}
Given any $j$-th unlabeled target data $\mathbf{x}_j^{trg}$ in $\mathcal{X}_{trg}$, we first employ the target model $\phi_{trg}$ to generate the pseudo-label $y_j^p=\arg \max _c p_{\phi_{trg}}(y_c|\mathbf{x}_j^{trg})$, where $p_{\phi_{trg}}(y_c|\mathbf{x}_j^{trg})=[p(y|\mathbf{x}_j^{trg} ; \phi_{trg})]_c$ denotes the probability corresponding to the $c$-th class in the output logits $[p(y | \mathbf{x}_j^{trg} ; \phi_{trg})]$ of $\phi_{trg}$. Existing research demonstrates that a small subset of pseudo-labels is trustworthy, while the rest are intrinsically unreliable~\cite{liang2021source,wang2021source,ding2023proxymix,yang2022divide,yang2023casting,chu2023adversarial,zhang2022divide}. The uncertainty can be measured by entropy to distinguish reliable and unreliable pseudo-labels~\cite{liang2021source,wang2021source,ye2021source,liu2021source}. Therefore, we compute the entropy $H_j^{trg}$ of the target model’s prediction $[p(y|\mathbf{x}^{trg}; \phi_{trg})]$ and divide $\mathcal{X}_{trg}$ into trust set $\mathcal{V}$ and non-trust set $\mathcal{U}$ using a threshold $E$. For sample $\mathbf{x}_k^{trg}$ with $H_k^{trg} \leq E$, we consider its pseudo label $y_k^p$ reliable and include $(\mathbf{x}_k^{trg},y_k^p)$ in $\mathcal{V}$. Otherwise, we solely include the sample in $\mathcal{U}$. Ultimately we obtain the trust set $\mathcal{V}=\{(\mathbf{x}_k^v,y_k^p)\}_{k=1}^{N_v}$ of $N_v$ samples and non-trust set $\mathcal{U}=\{(\mathbf{x}_l^u)\}_{l=1}^{N_u}$ of $N_u$ samples.

\subsection{Manipulation of Non-trust Set}\label{nottrustmanipulation}

In this section, we develop a diffusion-based manipulation strategy to further exploit the target-domain information inherently encapsulated in the unreliable pseudo-labels for non-trust samples $\mathbf{x}_l^u \in \mathcal{U}$. For each $\mathbf{x}_l^u \in \mathcal{U}$, we first uniformly assign a category label $\hat{y}_l$ to mitigate potential class imbalance.
\begin{equation}
\hat{y}_l = \left(l \bmod \left\lfloor |\mathcal{U}|/C \right\rfloor \right), \quad l \in \{1,2,...,\left\lfloor |\mathcal{U}|/C \right\rfloor \times C\},
\end{equation}
where $C$ is the total number of classes. Note that we discard the residual samples with $ \left\lfloor|\mathcal{U}|/C\right\rfloor \times C<l 
\leq |\mathcal{U}|$ for class balance. Subsequently, we achieve two objectives via a pre-trained diffusion model at the same time, \emph{i.e.}, i) semantic transformation of $\mathbf{x}_l^u$ toward the specified class $\hat{y}_l$ to convert $\hat{y}_l$ into an effective supervisory signal, and ii) preservation of the target-domain features. The manipulated sample $\tilde{\mathbf{x}}_l^{u}$ maintains fidelity to the target distribution while exhibiting substantially improved class certainty to allow for converting problematic non-trust samples into useful training instances.

To this end, our diffusion-based manipulation strategy consists of three key components, \emph{i.e.}, i) \textbf{Target-guided Initialization} that extracts target domain guidance from $\mathbf{x}_l^u$ to form an effective starting point for the diffusion denoising process, ii) \textbf{Semantic Feature Injection} that ensures the designated class $\hat{y}_l$ for generated samples during denoising, and \textbf{Domain-specific Feature Preservation} that maintains the generated samples within the target distribution, as elaborated below.

\textbf{Target-guided Initialization.} The sampling starting point $x_T$ of diffusion models, particularly its low-frequency components, has been proven to significantly influence the generated image~\cite{wu2023freeinit,qiu2023freenoise,koo2024flexiedit,xu2025stylessp,everaert2024exploiting}. During inference, the low-frequency components of the generated image and starting point for sampling $x_T$ remain strongly correlated and diffusion models exploit signal leakage from these low-frequency components for image generation~\cite{everaert2024exploiting}. To generate a novel sample from $\mathbf{x}_l^u$ that preserves target domain characteristics while conforming to the newly-assigned category $\hat{y}_l$ of $\mathbf{x}_l^u$, we propose to incorporate the inherent domain-specific features of $\mathbf{x}_l^u$ into the starting point for sampling in diffusion models.

In domain adaptation and generalization, domain-specific features are typically associated with the low-frequency components of $\mathbf{x}_l^u$. Furthermore, to prevent the potential semantic leakage from the high-frequency component of $\mathbf{x}_l^u$, we extract the high-frequency component $\mathcal{F}_{I_G}^H$ from semantically neutral random Gaussian noise $I_G$ and the low-frequency component $\mathcal{F}_{\mathbf{x}_l^u}^L$ from the input image $\mathbf{x}_l^u$ via Fast Fourier Transform ($\mathcal{FFT}$).
\begin{equation}
\mathcal{F}_{\mathbf{x}_l^u}^L =\mathcal{F} \mathcal{F} \mathcal{T}\left(\mathbf{x}_l^u\right) \odot \mathcal{H}, \quad
\mathcal{F}_{I_G}^H =\mathcal{F} \mathcal{F} \mathcal{T}(I_G) \odot(1-\mathcal{H}),
\end{equation}
where $\mathcal{H}$ is a low-pass filter. $\mathcal{F}_{\mathbf{x}_l^u}^L$ and $\mathcal{F}_{I_G}^H$ are combined and inversely transformed via inverse FFT ($\mathcal{IFFT}$) to produce a semantically neutral target-domain pseudo-image $\tilde{\mathbf{x}}_l^u$. 
\begin{equation}
\tilde{\mathbf{x}}_l^u  =\mathcal{I F FT}\left(\mathcal{F}_{\mathbf{x}_l^u}^L+\mathcal{F}_{I_G}^H \right),
\end{equation}
$\tilde{\mathbf{x}}_l^u$ is first encoded into the latent space via an encoder $\mathcal{E}$, and then subjected to a $T$-step DDPM forward process to add Gaussian noise. The noisy latent $\mathbf{z}_T$ is used as the starting point for sampling. 
\begin{equation}\label{initial}
\hat{\mathbf{z}}_0 = \mathcal{E}(\tilde{\mathbf{x}}_l^u), \quad 
    \mathbf{z}_T =  \sqrt{\alpha_T} \hat{\mathbf{z}}_0 + \sqrt{1-\alpha_T} \boldsymbol{\epsilon}, \quad \boldsymbol{\epsilon} \sim \mathcal{N}(\mathbf{0}, \mathbf{I}).
\end{equation}

\textbf{Semantic Feature Injection.} For our task, the sampling starting point $\mathbf{z}_T$ derived from~\eqref{initial} may inherently lack sufficient semantic relevance to $\hat{y}_l$ due to the following reasons. First, we construct the high-frequency components of $\mathbf{z}_T$ using a semantically neutral Gaussian noise image, which carries no $\hat{y}_l$-related information. Secondly, although we isolate the high-frequency components of $\mathbf{x}_l^u$, weak semantic leakage from $\mathbf{x}_l^u$ may persist, potentially conflicting with $\hat{y}_l$. Consequently, we may fail to semantically transform $\mathbf{x}_l^u$ to $\hat{y}_l$ even with a large guidance scale $\gamma_1$ according to~\cite{bai2024zigzag}. To address this, we present semantic feature injection as below.

During denoising, at each timestep 
$t$, we adopt a zigzag self-reflection operation following~\cite{bai2024zigzag}. We first denoise the latent $\mathbf{z}_t$ via the latent diffusion model to obtain $\mathbf{z}_{t-1}$, and then yield the refined latent $\tilde{\mathbf{z}}_t$ by injecting $\hat{y}_l$-related semantic information into $\mathbf{z}_{t-1}$ with DDIM inversion~\cite{song2020denoising}.
\begin{equation}\label{inversionstep}
\begin{aligned}
\tilde{\mathbf{z}}_t =\sqrt{\frac{\alpha_t}{\alpha_{t-1}}} \mathbf{z}_{t-1}+&\sqrt{\alpha_t}\left(\sqrt{\frac{1}{\alpha_t}-1}-\sqrt{\frac{1}{\alpha_{t-1}}-1}\right) \tilde{\epsilon}_\theta\left(\mathbf{z}_{t-1},\hat{y}_l,t-1\right) \\
\tilde{\epsilon}_\theta\left(\mathbf{z}_{t-1},\hat{y}_l,t-1\right)=&\left(1+\gamma_2\right) \epsilon_\theta\left(\mathbf{z}_{t-1}, \hat{y}_l, t-1\right)-\gamma_2 \epsilon_\theta\left(\mathbf{z}_{t-1}, \varnothing, t-1\right),
\end{aligned}
\end{equation}
where $\gamma_2$ is the inversion guidance scale. According to \eqref{inversionstep}, semantic alignment with $\hat{y}_l$ is considered for the latents throughout the sampling trajectory. However, since DDIM inversion could introduce unrelated domain features, the latents could deviate from the target distribution when accumulating $\hat{y}_l$-aligned semantic information.
To address this, we selectively extract the high-frequency components carrying the accumulated semantic information and discard the low-frequency components harboring domain artifacts from $\tilde{\mathbf{z}}_t$ rather than directly leveraging $\tilde{\mathbf{z}}_t$. 
\begin{equation}\label{inversionfft}
\mathcal{F}_{\tilde{\mathbf{z}}_t}^H =\mathcal{F} \mathcal{F} \mathcal{T}(\tilde{\mathbf{z}}_t) \odot(1-\mathcal{H}).
\end{equation}
$\mathcal{F}_{\tilde{\mathbf{z}}_t}^H$ is used as high-frequency semantics to aggregate with target domain specific features.

\textbf{Domain-specific Feature Preservation.} To better align with the target distribution, we combine the high-frequency semantic features $\mathcal{F}_{\tilde{\mathbf{z}}_t}^H$ in the latents by DDIM inversion  in~\eqref{inversionfft} with the target domain specific features at each denoising timestep $t$. The domain-specific features are primarily encoded in the low-frequency components of samples from the target domain. To adapt to the time-varying noise level in $\mathbf{z}_t$, we perturb the clean latent $\hat{\mathbf{z}}_0$ in~\eqref{initial} via the DDPM forward process with $t$-step Gaussian noise to generate $\hat{\mathbf{z}}_{0,t} =  \sqrt{\alpha_t} \hat{\mathbf{z}}_0 + \sqrt{1-\alpha_t} \boldsymbol{\epsilon}$ for timestep $t$. The low-frequency domain-specific features $\mathcal{F}_{\hat{\mathbf{z}}_{0,t}}^L$ are extracted from $\hat{\mathbf{z}}_{0,t}$ and combined with $\mathcal{F}_{\tilde{\mathbf{z}}_t}^H$ to obtain enhanced latent $\tilde{\mathbf{z}}^\prime_t$ that simultaneously preserves $\hat{y}_l$-aligned high-frequency semantics and embeds target-domain low-frequency features to produce $\mathbf{z}_{t-1}$.

\begin{equation}
\begin{aligned}
\tilde{\mathbf{z}}^\prime_t =\mathcal{I F FT}\left(\mathcal{F}_{\hat{\mathbf{z}_{0,t}}}^L+\mathcal{F}_{\tilde{\mathbf{z}}_t}^H \right),\quad \mathcal{F}_{\hat{\mathbf{z}}_{0,t}}^L=\mathcal{F} \mathcal{F} \mathcal{T}\left(\hat{\mathbf{z}}_{0,t}\right) \odot \mathcal{H}.
\end{aligned}
\end{equation}

\subsection{Progressive Refinement Mechanism}\label{updatemechanism}

We design a progressive refinement mechanism to iteratively refine the pseudo-target domain for $R$ iterations to further optimize the target model. When optimized solely on a fixed pseudo-target domain, the target model could be affected by the trust set $\mathcal{V}$ inevitably contains residual label noise due to imperfect pseudo-labeling, and the manipulated non-trust set $\mathcal{U}^m$ gradually accumulates domain discrepancy in sample generation. Therefore, for any $r$-th ($r=1,\cdots,R$) refinement iteration, we update the trust set $\mathcal{V}^{r}$ to correct inaccurate pseudo-labels and reduce the size of the non-trust set $\mathcal{U}^{r}$ for decreasing domain discrepancy. The target model $\phi_{trg}^{r}$ re-partitions the target data into updated trust set $\mathcal{V}^{(r+1)}$ and non-trust set $\mathcal{U}^{(r+1)}$. $\mathcal{U}^{(r+1)}$ is then further manipulated according to Section~\ref{nottrustmanipulation} to generate $\mathcal{U}^{m,(r+1)}$ for constructing the refined pseudo-target domain $\mathcal{D}_p^{(r+1)} = \mathcal{V}^{(r+1)} \cup \mathcal{U}^{m,(r+1)}$. The updated target model $\phi_{trg}^{(r+1)}$ is obtained by fine-tuning $\theta_v^{(r)}$ on $\mathcal{D}_p^{(r+1)}$. We empirically find in Figure~\ref{fig:ab1} that, during progressive refinement,  $\mathcal{V}^{(r+1)}$ provides more accurate pseudo-labels than $\mathcal{V}^{(r)}$ with $|\mathcal{V}^{(r+1)}| > |\mathcal{V}^{(r)}|$ such that $|\mathcal{U}^{m,(r+1)}| < |\mathcal{U}^{m,(r)}|$ to reduce the size of manipulated non-trust set and decrease the domain discrepancy in the pseudo-target domain. Compared with $\phi_{trg}^{(r)}$, $\phi_{trg}^{(r+1)}$ can better approximate the real target distribution and finally achieve enhanced performance.  

\section{Experiments}

\begin{table}[!t]
\renewcommand{\baselinestretch}{1.0}
\renewcommand\arraystretch{1.0}
\setlength{\tabcolsep}{3pt}
\centering
\caption{\textbf{Office-31} results (\%) with ResNet-50. Methods with top three performance in each column are highlighted in red, orange, and yellow.
}\label{tab:oc}
\scriptsize
\begin{tabular}{@{} l l | c c c c c c c@{}}
        \toprule
        Method &Venue   &A$\to$D &A$\to$W &D$\to$A &D$\to$W &W$\to$A &W$\to$D &Avg.\\
        \toprule
        \multicolumn{9}{c}{Baseline method} \\
        \midrule
        Source        &--     &79.7 &77.6 &65.5 &97.9 &63.8 &99.8 &80.7 \\
        \midrule
        \multicolumn{9}{c}{Generation-based method} \\
        \midrule 
        CPGA~\cite{cpga}        &IJCAI21     &94.4 &94.1 &76.0 &98.4 &76.6 &99.8 &89.9 \\
        ASOGE~\cite{asoge}        &TCSVT23     &95.6 &94.1 &74.3 &98.1 &74.2 &99.7 &89.3 \\
        ISFDA~\cite{improvedsfda}        &CVPR24     &95.3 &94.2 &76.4 &98.3 &77.5 &\besttwo{99.9} &90.3 \\
        DM-SFDA~\cite{dmsfda}  &- &\best{97.7} &\best{99.0} &\besttwo{82.7} &\best{99.3} &\besttwo{83.5} &\best{100.0} &\besttwo{93.7} \\
        \midrule 
        \multicolumn{9}{c}{None-generation method} \\
        \midrule
        SHOT~\cite{shot}    &ICML20    &93.7 &91.1 &74.2 &98.2 &74.6 &\best{100.} &88.6 \\ 
        NRC~\cite{nrc}     &NIPS21    &96.0 &90.8 &75.3 &\bestthree{99.0} &75.0 &\best{100.} &89.4 \\
        GKD~\cite{gkd}     &IROS21    &94.6 &91.6 &75.1 &98.7 &75.1 &\best{100.} &89.2 \\
        HCL~\cite{hcl}     &NIPS21    &94.7 &92.5 &75.9 &98.2 &77.7 &\best{100.} &89.8 \\
        AaD~\cite{ada}     &NIPS22    &96.4 &92.1 &75.0 &\besttwo{99.1} &76.5 &\best{100.} &89.9 \\ 
        AdaCon~\cite{adacon}  &CVPR22    &87.7 &83.1 &73.7 &91.3 &77.6 &72.8 &81.0 \\
        CoWA~\cite{cowa}    &ICML22    &94.4 &\bestthree{95.2} &76.2 &98.5 &77.6 &\bestthree{99.8} &90.3 \\
        ELR~\cite{elr}     &ICLR23    &93.8 &93.3 &76.2 &98.0 &76.9 &\best{100.} &89.6 \\
        PLUE~\cite{plue}    &CVPR23    &89.2 &88.4 &72.8 &97.1 &69.6 &97.9 &85.8 \\
        CPD~\cite{cpd}     &PR24         &96.6 &94.2 &77.3 &98.2 &78.3 &\best{100.} &\bestthree{90.8}\\
        TPDS~\cite{tpds}    &IJCV24       &\bestthree{97.1} &94.5 &75.7 &98.7 &75.5 &\bestthree{99.8} &90.2 \\
        DIFO~\cite{ada}  &CVPR24    &93.6 &92.1 &78.5 &95.7 &78.8 &97.0 &89.3 \\
        ProDe~\cite{prode} &ICLR25    &94.4 &92.1 &\bestthree{79.8} &95.6 &\bestthree{79.0} &98.6 &89.9 \\
        \midrule
        \textbf{DPTM(ours)} &--    &\besttwo{97.2} &\besttwo{95.3} &\best{92.0} & 98.7 &\best{91.7} &\best{100.} &\best{95.8} \\
\bottomrule
\end{tabular}
\end{table}

\subsection{Experimental Settings}

\textbf{Datasets.} We adopt four standard domain adaptation benchmarks of different scales for evaluations, including the small-scale Office-31 dataset~\cite{office31}, the medium-scale Office-Home dataset~\cite{officehome}, and two large-scale datasets (\emph{i.e.}, VisDA~\cite{visda} and DomainNet-126~\cite{domainnet}). Refer to the supplementary material for complete dataset statistics and domain configurations.

\textbf{Comparative Methods.}
We compare with 21 existing methods from three distinct groups: i) the baseline results from the source model, ii) \textbf{generation-based SFDA methods} CPGA~\cite{cpga}, ASOGE~\cite{asoge}, ISFDA~\cite{improvedsfda}, PS~\cite{ps}, DATUM~\cite{datum}, and DM-SFDA~\cite{dmsfda}, and iii) \textbf{non-generation SFDA methods} including current state-of-the-art SFDA methods SHOT~\cite{shot}, NRC~\cite{nrc}, GKD~\cite{gkd}, HCL~\cite{hcl}, AaD~\cite{ada}, AdaCon~\cite{adacon}, CoWA~\cite{cowa}, SCLM~\cite{sclm}, ELR~\cite{elr}, PLUE~\cite{plue}, CRS~\cite{crs}, CPD~\cite{cpd}, TPDS~\cite{tpds}, DIFO~\cite{difo}, and ProDe~\cite{prode}.

\textbf{Implementation Details.} We employ stable-diffusion v1-5~\cite{rombach2022high} as the diffusion model to generate 512$\times$512 images with 20 denoising steps. $\gamma_1=5.5$ in~\eqref{denoising} and $\gamma_2=0$ in~\eqref{inversionstep}. We set the threshold $E$ to 0.01, and the total refinement iteration count $R$ to 10. Note that setting $E$ and $R$ to other values may obtain superior performance. 
For the adaptation model, we employ ResNet-50 for Office-31~\cite{office31}, Office-Home~\cite{officehome} and 
DomainNet-126~\cite{domainnet}, and ResNet-101 for VisDA~\cite{visda}. We train for 20K iterations with the batch size of 128 and learning rate of 3\emph{e}-3 for large-scale DomainNet-126~\cite{domainnet} and VisDA~\cite{visda}, and 15K iterations with the batch size of 32 and learning rate of 1\emph{e}-3 for Office-31 and Office-Home. Weight decay is set to 5\emph{e}-4 for all the datasets.

\subsection{Main Results}
\textbf{Evaluations on Office-31.} 
Table~\ref{tab:oc} shows that our method is superior to generation-based SFDA methods on Office-31 and outperforms the best generation-based SFDA method DM-SFDA~\cite{dmsfda} on average across all the DA tasks. Compared with non-generation methods, our method outperforms the best non-generation methods in all tasks except D$\to$W, delivering an average accuracy gain of 5\%. Notably, our method achieves significant improvements on challenging adaptation tasks: 9.3\% on D$\to$A and 8.2\% on W$\to$A. These results validate the effectiveness of our method.

\textbf{Evaluations on Office-Home and Visda.}
Table~\ref{tab:oh_vi} shows that our method significantly outperforms existing SFDA methods on Office-Home and VisDA. On Office-Home, we achieve an average accuracy gain of 11.7\% over the best generation-based SFDA method DM-SFDA~\cite{dmsfda} and 10.1\% over the current SOTA method ProDe~\cite{prode} across all domain adaptation tasks. Remarkably, our method outperforms ProDe by 22.7\%, 21.0\%, and 21.6\% on challenging Ar$\to$Cl, Pr$\to$Cl, and Rw$\to$Cl tasks where existing methods usually perform poorly. On VisDA, our method achieves an average accuracy gain of 8.5\% over ISFDA~\cite{improvedsfda} and 8.2\% ProDe~\cite{prode} (see the supplementary material for details). These results strongly validate the effectiveness of our method in difficult domain adaptation scenarios.

\begin{table}[!t]
\renewcommand\tabcolsep{0.6pt}
\renewcommand\arraystretch{1.0}
\centering
\caption{Results~(\%) on \textbf{Office-Home} and \textbf{VisDA}. \textbf{Office-Home} is evaluated with ResNet-50, and \textbf{VisDA} is evaluated with ResNet-101. The top three performances in each column are highlighted in red, orange, and yellow, respectively.}\label{tab:oh_vi}
\scriptsize
\begin{tabular}{@{} l l |  c c c c c c c c c c c c c |c@{}}
        \toprule
        \multirow{2}{*}{Method} &\multirow{2}{*}{Venue} 
        &\multicolumn{13}{c}{\textbf{Office-Home}} &\multicolumn{1}{c}{\textbf{VisDA}}\\
        & &Ar$\to$Cl &Ar$\to$Pr &Ar$\to$Rw
        &Cl$\to$Ar &Cl$\to$Pr &Cl$\to$Rw    
        &Pr$\to$Ar &Pr$\to$Cl &Pr$\to$Rw  
        &Rw$\to$Ar &Rw$\to$Cl &Rw$\to$Pr &Avg. &Sy$\to$Re\\
        \midrule
        \multicolumn{16}{c}{Baseline method} \\
        \midrule
        Source        &--  
        &50.1 &67.9 &74.4 &55.2 &65.2 &67.2 &53.4 &44.5 &74.1 &64.2 &51.5 &78.7 &62.2  &63.5\\
        \midrule
        \multicolumn{16}{c}{Generation-based method} \\
        \midrule
        CPGA~\cite{cpga}        &IJCAI21  
        &59.3 &78.1 &79.8 &65.4 &75.5 &76.4 &65.7 &58.0 &81.0 &72.0 &64.4 &83.3 &71.6 &86.0\\
        ASOGE~\cite{asoge}       &TCSVT23  
        &59.1 &78.4 &81.0 &67.7 &78.4 &77.5 &65.8 &57.2 &80.2 &72.7 &60.7 &83.3 &71.8 &83.2
        \\
        ISFDA~\cite{improvedsfda}        &CVPR24  
        &60.7 &78.9 &82.0 &69.9 &79.5 &79.7 &67.1 &58.8 &82.3 &74.2 &61.3 &86.4 &73.4 &88.4\\
        PS~\cite{ps}        &ML24  
        &57.8 &77.3 &81.2 &68.4 &76.9 &78.1 &67.8 &57.3 &82.1 &75.2 &59.1 &83.4 &72.1 &84.1\\
        DATUM~\cite{datum} &CVPR23 &55.3 &76.8 &79.3 &65.1 &77.7 &78.6 &62.4 &52.1 &79.7 &66.6 &55.9 &80.5 &69.2 &-- \\
        DM-SFDA~\cite{dmsfda} &-- &\besttwo{68.5} &\bestthree{89.6} &83.3 &70.0 &85.8 &\bestthree{87.4} &71.3 &\besttwo{69.6} &\bestthree{88.2} &77.8 &\besttwo{68.5} &\bestthree{88.7} &\bestthree{79.5} &86.3 \\
        \midrule
        \multicolumn{16}{c}{None-generation method} \\
        \midrule
        SHOT~\cite{shot}     &ICML20    &56.7 &77.9 &80.6 &68.0 &78.0 &79.4 &67.9 &54.5 &82.3 &74.2 &58.6 &84.5 &71.9 &82.7\\
        NRC~\cite{nrc}      &NIPS21    &57.7 &80.3 &82.0 &68.1 &79.8 &78.6 &65.3 &56.4 &83.0 &71.0 &58.6 &85.6 &72.2 &85.9\\
        GKD~\cite{gkd}      &IROS21   &56.5 &78.2 &81.8 &68.7 &78.9 &79.1 &67.6 &54.8 &82.6 &74.4 &58.5 &84.8 &72.2 &83.0\\ 
        AaD~\cite{ada}      &NIPS22    &59.3 &79.3 &82.1 &68.9 &79.8 &79.5 &67.2 &57.4 &83.1 &72.1 &58.5 &85.4 &72.7 &88.0\\ 
        AdaCon~\cite{adacon}   &CVPR22    &47.2 &75.1 &75.5 &60.7 &73.3 &73.2 &60.2 &45.2 &76.6 &65.6 &48.3 &79.1 &65.0 &86.8\\ 
        CoWA~\cite{cowa}     &ICML22    &56.9 &78.4 &81.0 &69.1 &80.0 &79.9 &67.7  &57.2 &82.4 &72.8 &60.5 &84.5 &72.5 &86.9\\
        SCLM~\cite{sclm}     &NN22    
        &58.2 &80.3 &81.5 &69.3 &79.0 &80.7 &69.0 &56.8 &82.7 &74.7 &60.6 &85.0 &73.0 &85.3\\
        ELR~\cite{elr}      &ICLR23    &58.4 &78.7 &81.5 &69.2 &79.5 &79.3 &66.3 &58.0 &82.6 &73.4 &59.8 &85.1 &72.6 &85.8\\ 
        PLUE~\cite{plue}     &CVPR23    &49.1 &73.5 &78.2 &62.9 &73.5 &74.5 &62.2 &48.3 &78.6 &68.6 &51.8 &81.5 &66.9 &88.3\\ 
        CPD~\cite{cpd}      &PR24      &59.1 &79.0 &82.4 &68.5 &79.7 &79.5 &67.9 &57.9 &82.8 &73.8 &61.2 &84.6 &73.0 &85.8\\ 
        TPDS~\cite{tpds}     &IJCV24    &59.3 &80.3 &82.1 &70.6 &79.4 &80.9 &69.8 &56.8 &82.1 &74.5 &61.2 &85.3 &73.5 &87.6\\ 
        DIFO~\cite{difo}  &CVPR24     & 62.6 &  87.5 &  \bestthree{87.1} &  \bestthree{79.5} &  \bestthree{87.9} &  \bestthree{87.4} &  \bestthree{78.3} &  63.4 &  88.1 &  \bestthree{80.0} &  63.3 &  87.7 &  79.4 &  \bestthree{88.6}\\ 
        ProDe~\cite{prode}   &ICLR25    & \bestthree{64.0} & \besttwo{90.0} & \besttwo{88.3} & \besttwo{81.1} & \besttwo{90.1} & \besttwo{88.6} & \besttwo{79.8} & \bestthree{65.4} & \besttwo{89.0} & \besttwo{80.9} & \bestthree{65.5} & \besttwo{90.2} & \besttwo{81.1} & \besttwo{88.7} \\
        \midrule
         \textbf{DPTM(ours)}   &--    
    &\best{86.7} 	&\best{94.2} 	&\best{92.8} 	&\best{91.5} 	&\best{94.0} 	&\best{92.6} 	&\best{90.6} 	&\best{86.4} 	&\best{92.8} 	&\best{90.5} 	&\best{87.1} 	&\best{94.7} 	&\best{91.2}   &\best{97.6}
     \\
    \bottomrule
    \end{tabular}
\end{table}
\begin{table}[t]
\renewcommand\tabcolsep{4pt}
\renewcommand\arraystretch{1.0}
\centering
\caption{Results~(\%) on \textbf{DomainNet-126} evaluated with ResNet-50. The top three performances in each column are highlighted in red, orange, and yellow, respectively.}\label{tab:dn}
\scriptsize
\begin{tabular}{@{} l l |  c c c c c c c c c c c c c@{}}
\toprule
        \multirow{2}{*}{Method} &\multirow{2}{*}{Venue} 
        &\multicolumn{13}{c}{\textbf{DomainNet-126}} \\
        & &C$\to$P &C$\to$R &C$\to$S
        &P$\to$C &P$\to$R &P$\to$S    
        &R$\to$C &R$\to$P &R$\to$S  
        &S$\to$C &S$\to$P &S$\to$R &Avg. \\
        \midrule
        \multicolumn{15}{c}{Baseline method} \\
        \midrule
        Source        &--  
        &47.5 &59.8 &48.6 &51.0 &75.3 &47.8 &57.5 &61.1 &48.6 &63.5 &56.2 &59.5 &56.4 \\
        \midrule
        \multicolumn{15}{c}{Generation-based method} \\
        \midrule
        CPGA~\cite{cpga}        &IJCAI21
        &61.2 &76.7 &59.6 &64.5 &81.3 &61.0 &68.6 &69.5 &65.9 &66.9 &60.2 &75.1 &67.6 \\
        \midrule
        \multicolumn{15}{c}{None-generation method} \\
        \midrule
        SHOT~\cite{shot}     &ICML20    &63.5 &78.2 &59.5 &67.9 &81.3 &61.7 &67.7 &67.6 &57.8 &70.2 &64.0 &78.0 &68.1 \\
        NRC~\cite{nrc}      &NIPS21    &62.6 &77.1 &58.3 &62.9 &81.3 &60.7 &64.7 &69.4 &58.7 &69.4 &65.8 &78.7 &67.5 \\
        GKD~\cite{gkd}      &IROS21   &61.4 &77.4 &60.3 &69.6 &81.4 &63.2 &68.3 &68.4 &59.5 &71.5 &65.2 &77.6 &68.7 \\ 
        AdaCon~\cite{adacon}   &CVPR22    &60.8 &74.8 &55.9 &62.2 &78.3 &58.2 &63.1 &68.1 &55.6 &67.1 &66.0 &75.4 &65.4 \\ 
        CoWA~\cite{cowa}     &ICML22    &64.6 &80.6 &60.6 &66.2 &79.8 &60.8 &69.0 &67.2 &60.0 &69.0 &65.8 &79.9 &68.6 \\
        PLUE~\cite{plue}     &CVPR23    &59.8 &74.0 &56.0 &61.6 &78.5 &57.9 &61.6 &65.9 &53.8 &67.5 &64.3 &76.0 &64.7\\ 
        TPDS~\cite{tpds}     &IJCV24    &62.9 &77.1 &59.8 &65.6 &79.0 &61.5 &66.4 &67.0 &58.2 &68.6 &64.3 &75.3 &67.1 \\ 
        DIFO~\cite{difo}  &CVPR24     &\bestthree{73.8} & \bestthree{89.0} & \bestthree{69.4} & \bestthree{74.0} & \bestthree{88.7} & \bestthree{70.1} & \bestthree{74.8} & \bestthree{74.6} & \bestthree{69.6} & \bestthree{74.7} & \bestthree{74.3} & \bestthree{88.0} & \bestthree{76.7}\\ 
        ProDe~\cite{prode}   &ICLR25    & \besttwo{79.3} &\best{91.0} & \besttwo{75.3} & \besttwo{80.0} &\best{90.9} & \besttwo{75.6} & \besttwo{80.4} & \besttwo{78.9} & \besttwo{75.4} & \besttwo{80.4} & \besttwo{79.2} &\best{91.0} & \besttwo{81.5} \\
        \midrule
        \textbf{DPTM(ours)}   &--    
    &\best{85.6} 	& \besttwo{90.9} 	&\best{80.0} 	&\best{85.1} 	&\besttwo{90.7}	&\best{79.0} 	&\best{85.2} 	&\best{85.4} 	&\best{78.1} 	&\best{86.1} 	&\best{85.4} 	&\besttwo{90.9} 	&\best{85.2}  
     \\
    \bottomrule
\end{tabular}
\end{table}

\textbf{Evaluations on DomainNet-126.} Our method achieves a 17.6\% higher average accuracy than the generation-based CPGA~\cite{cpga} and surpasses current SOTA ProDe~\cite{prode} by 3.7\% on DomainNet-126. It significantly outperforms CPGA~\cite{cpga} across all domain adaptation tasks and exceeds ProDe~\cite{prode} in most tasks, with only minor performance gaps in three DA scenarios.

\subsection{Ablation Studies}\label{moreanalysis}
We conduct ablation studies mainly on the Office-Home dataset. More ablation studies can be found in the supplementary materials.

\begin{table}[!t]
\renewcommand{\baselinestretch}{1.0}
\renewcommand\arraystretch{1.0}
\renewcommand\tabcolsep{0.5pt}
\centering
\caption{Ablation study results~(\%) on Different LDMs evaluated with $R=3$, $E=0.001$.}\label{absd}
\footnotesize 
\begin{tabular}{@{}l  |  c c c c c c c c c c c c c@{}}
\toprule
LDM &Ar$\to$Cl &Ar$\to$Pr &Ar$\to$Rw &Cl$\to$Ar &Cl$\to$Pr &Cl$\to$Rw &Pr$\to$Ar &Pr$\to$Cl &Pr$\to$Rw &Rw$\to$Ar &Rw$\to$Cl &Rw$\to$Pr &Avg. \\
\midrule
SDXL    &\textbf{69.4} &82.2 &82.8 &69.4 &82.6 &82.2 &\textbf{65.8} &\textbf{67.0} &82.4 &\textbf{71.1} &\textbf{67.5} &84.2 &\textbf{75.6}   \\
SD15    &67.0 &\textbf{83.6} &\textbf{83.9} &\textbf{70.6} &\textbf{84.3} &\textbf{82.9} &65.6 &63.6 &\textbf{84.6} &69.6 &66.4 &\textbf{85.0}  &\textbf{75.6}   \\
\bottomrule
\end{tabular}
\end{table}
\begin{table}[!t]
\renewcommand{\baselinestretch}{1.0}
\renewcommand\arraystretch{1.0}
\renewcommand\tabcolsep{0.6pt}
\centering
\caption{Ablation study results~(\%) on Threshold $E$ evaluated with $R=10$.}\label{abe}
\footnotesize 
\begin{tabular}{@{}l  |  c c c c c c c c c c c c c@{}}
\toprule
$E$&Ar$\to$Cl &Ar$\to$Pr &Ar$\to$Rw &Cl$\to$Ar &Cl$\to$Pr &Cl$\to$Rw &Pr$\to$Ar &Pr$\to$Cl &Pr$\to$Rw &Rw$\to$Ar &Rw$\to$Cl &Rw$\to$Pr &Avg. \\
\midrule
$0.001$    &74.7 &85.9 &87.5 &75.0 &88.5 &85.6 &72.1 &73.3 &87.2 &75.2 &74.6 &88.8 &80.7   \\
$0.005$    &81.7 &92.6 &90.8 &85.5 &92.1 &88.2 &85.5 &81.2 &88.3 &82.1 &79.9 &92.1 &86.7   \\
$0.01$    &\textbf{86.7} &\textbf{94.2} &\textbf{92.8} &\textbf{91.5} &\textbf{94.0} &\textbf{92.6} &\textbf{90.6} &\textbf{86.4} &\textbf{92.8} &\textbf{90.5} &\textbf{87.1} &\textbf{94.7} &\textbf{91.2}   \\
\bottomrule
\end{tabular}
\end{table}

\begin{figure}[!t]
\renewcommand{\baselinestretch}{1.0}
\centering
\includegraphics[width=0.8\textwidth]{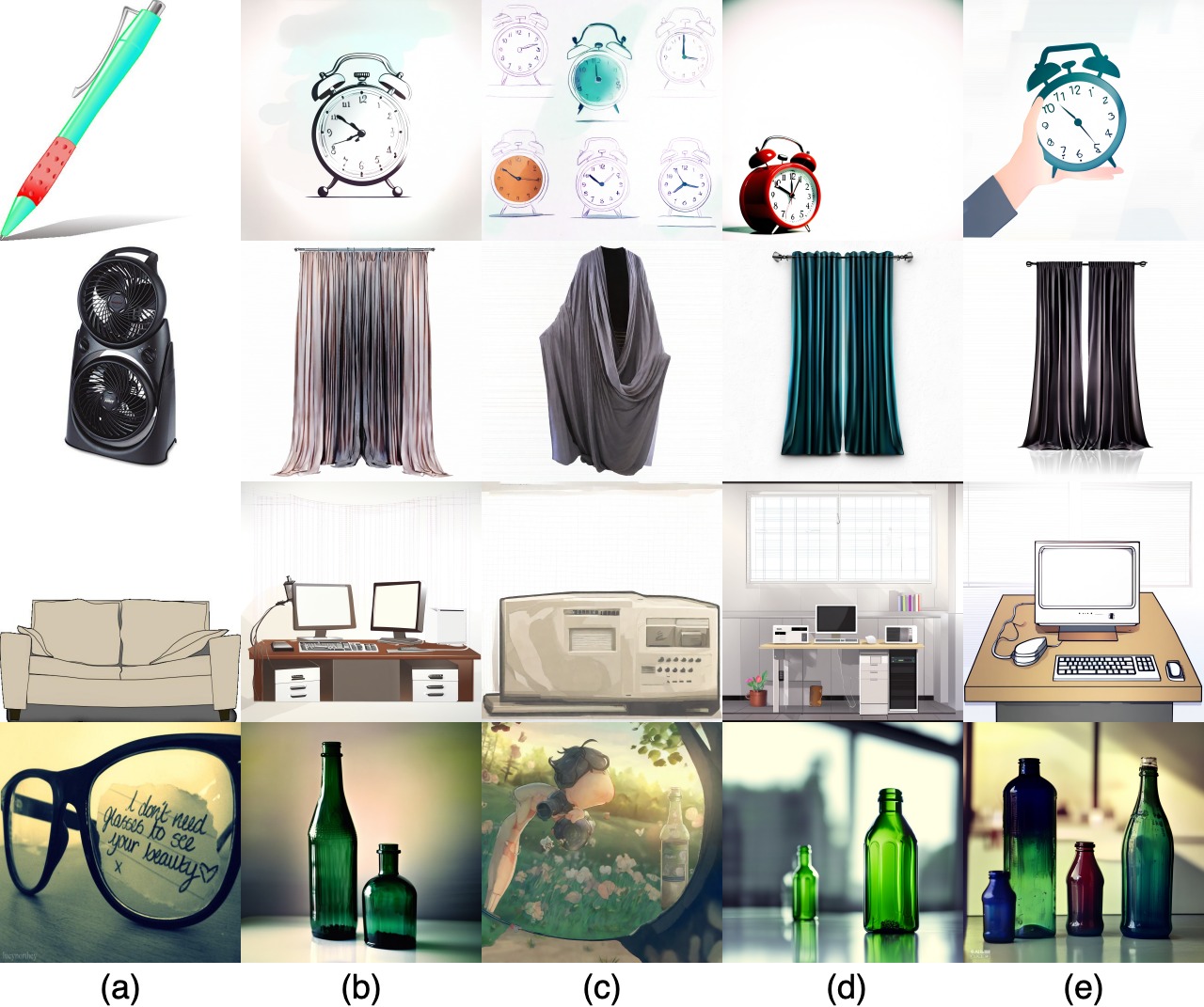} 
\caption{Ablation on Manipulation Mechanism of $\mathbf{x}_l^u$. Row: $\hat{y}_l=$ 'Alarm Clock', 'Curtains', 'Computer', 'Bottle', respectively. Column: (a) $\mathbf{x}_l^u$ (b) $\tilde{\mathbf{x}}_l^{u}$ w/o Target-guided Initialization (c) $\tilde{\mathbf{x}}_l^{u}$ w/o Semantic Feature Injection (d) $\tilde{\mathbf{x}}_l^{u}$ w/o Domain-specific Feature Preservation (e) $\tilde{\mathbf{x}}_l^{u}$ of our method.}\label{fig:vis1} 
\end{figure}

\textbf{Different Versions of Latent Diffusion Models.}
We conduct ablation studies using both Stable Diffusion v1.5 (SD15) and Stable Diffusion XL (SDXL)~\cite{sdxl}, with identical parameters ($E=0.001$ and $R=3$) except for output resolution - SDXL natively generates 1024$\times$1024 images while SD15 produces 512$\times$512 images due to their architectural differences. Table~\ref{absd} shows comparable performance, but SDXL's higher computational cost makes SD15 our preferred choice for implementation.

\textbf{Values of Threshold $E$.} We also conduct ablation on the Threshold $E=\{0.001,0.005,0.01\}$. Table~\ref{abe} shows that appropriately increasing $E$ may yield better results.

\textbf{Manipulation Mechanism of Non-trust Set.} As shown in Figure~\ref{fig:vis1}, our method's manipulated samples $\tilde{\mathbf{x}}_l^{u}$ exhibit the best semantic alignment with their assigned labels $\hat{y}_l=$ and the best preservation of target distribution characteristics, detailed in the supplementary material.

\subsection{Additional Analysis}
We provide additional analyses to further validate the effectiveness of our method. More analysis can be found in the supplementary materials.

\begin{figure}[!t]
\centering
\includegraphics[width=\textwidth]{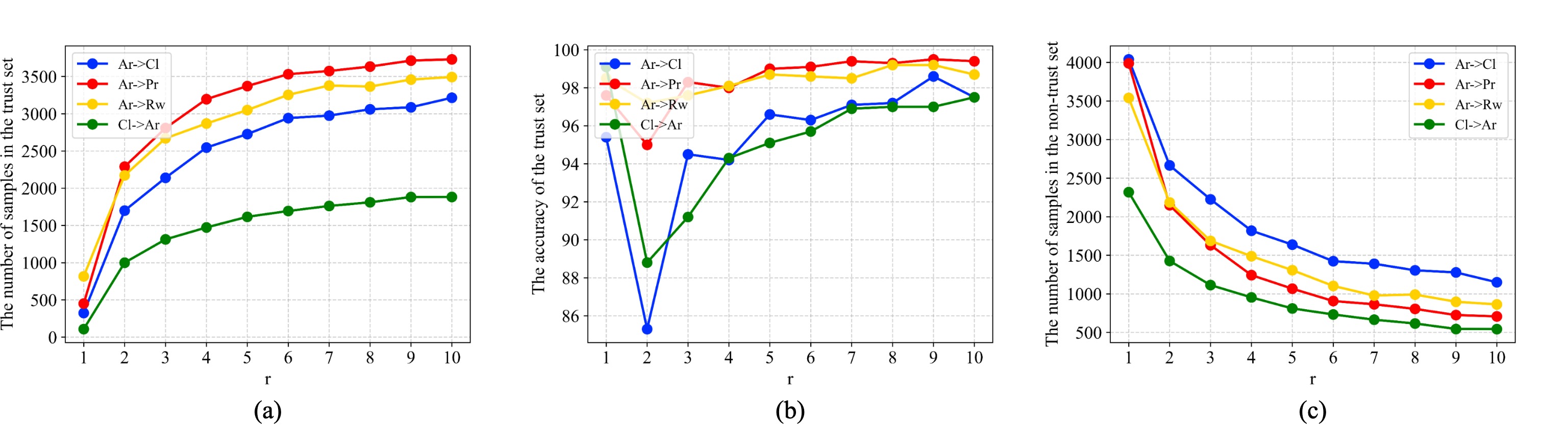} 
\caption{The relationship between $r$ versus: (a) The number of samples in the trust set. (b) The trust set accuracy. (c) The number of samples in the non-trust set.}\label{fig:ab1} 
\end{figure}

\textbf{Analysis on Progressive Refinement Mechanism.} 
For the performance of our method on the Office-Home dataset shown in Table \ref{tab:oh_vi}, we provide a detailed performance trajectory as $r$ increases from 1 to 10, in order to demonstrate the effectiveness of the proposed Progressive Refinement Mechanism. \textbf{Firstly}, we present experimental results for $r=\{0,2,4,6,8,10\}$ in Table~\ref{ab1r} and provide complete experimental results in the supplementary materials, where $r=0$ is equivalent to using only the source model. Table~\ref{ab1r} shows that the performance of the target model improves as $r$ increases. Specifically, as $r$ increases, the performance of the target model first improves rapidly and then growth becomes slow. \textbf{Secondly}, we select the first 4 DA tasks Ar$\to$Cl, Ar$\to$Pr, Ar$\to$Rw, and Cl$\to$Ar, and plot: the relationship between the number of samples in the trust set versus $r$, the trust set accuracy versus $r$, and the number of samples in the non-trust set versus $r$. As shown in Figure~\ref{fig:ab1}, two main conclusions can be drawn:
(1) The number of samples in the trust set increases significantly with $r$, while correspondingly, the non-trust set size decreases substantially. This indicates that the model progressively learns to make predictions with low uncertainty. (2) Overall, the trust set accuracy remains at a high level. Although relatively low across all four DA tasks at $r=2$, the accuracy shows significant recovery with increasing $r$, demonstrating our method's capability to progressively correct previous errors.

\textbf{Analysis on Trust and Non-trust Partition for Target Domain.} We provide more analysis on the proposed Trust and Non-trust Partition for the target Domain, detailed in the supplementary materials.

\textbf{Analysis on Manipulation of Non-trust Set.} We provide more analysis on the proposed Manipulation of Non-trust Set, detailed in the supplementary materials.

\begin{table}[!t]
\renewcommand{\baselinestretch}{1.0}
\renewcommand\arraystretch{1.0}
\renewcommand\tabcolsep{1pt}
\centering
\caption{Full results of the performance trajectory as $r$ grows from 1 to 10 on Office-Home evaluated with $E=0.01$.}\label{ab1r}
\footnotesize 
\begin{tabular}{@{}l  |  c c c c c c c c c c c c c@{}}
\toprule
$r$&Ar$\to$Cl &Ar$\to$Pr &Ar$\to$Rw
&Cl$\to$Ar &Cl$\to$Pr &Cl$\to$Rw  &Pr$\to$Ar &Pr$\to$Cl &Pr$\to$Rw 
&Rw$\to$Ar &Rw$\to$Cl &Rw$\to$Pr &Avg. \\
\midrule
$0$    &50.1 &67.9 &74.4 &55.2 &65.2 &67.2 &53.4 &44.5 &74.1 &64.2 &51.5 &78.7 &62.2 \\
$1$    &60.2  &80.9  &82.8  &67.8  &79.3  &80.4  &64.5  &57.1  &81.6  &69.1  &60.5  &83.4  &72.3  \\
$2$    &72.6 &87.2 &85.7 &73.5 &85.4 &84.4 &72.3 &69.7 &85.2 &76.7 &70.8 &87.7 &79.3  \\
$3$    &75.3  &89.7  &87.6  &77.7  &88.9  &86.4  &79.2  &76.1  &86.7  &80.1  &75.5  &90.2  &82.8  \\
$4$     &79.0 &91.8 &89.1 &81.0 &91.2 &88.3 &82.1 &79.1 &88.1 &82.7 &79.1 &91.7 &85.3   \\
$5$   &81.5  &92.5  &89.8  &83.9  &92.0  &89.4  &85.3  &81.5  &89.3  &84.5  &81.8  &92.8  &87.0  \\
$6$     &83.7 &92.9 &90.7 &86.1 &92.8 &90.2 &87.1 &83.1 &90.0 &87.4 &83.3 &93.3 &88.4   \\
$7$    &85.0  &93.7  &91.5  &87.5  &93.0  &91.2  &87.8  &84.4  &91.0  &88.6  &84.9  &93.9  &89.4  \\
$8$     &85.9 &94.2 &91.7 &88.7 &93.8 &91.9 &89.0 &85.1 &91.7 &89.7 &85.6 &94.2 &90.1   \\
$9$    &85.9  &94.1  &92.0  &89.8  &93.9  &92.3  &89.8  &85.6  &92.2  &90.2  &86.6  &94.4  &90.6  \\
$10$    &\textbf{86.7} &\textbf{94.2} &\textbf{92.8} &\textbf{91.5} &\textbf{94.0} &\textbf{92.6} &\textbf{90.6} &\textbf{86.4} &\textbf{92.8} &\textbf{90.5} &\textbf{87.1} &\textbf{94.7} &\textbf{91.2}   \\
    \bottomrule
    \end{tabular}
\end{table}

\section{Conclusion}
We propose DPTM, a novel generation-based framework that utilizes unlabeled target data as references to construct and progressively refine a pseudo-target domain via the latent diffusion model for Source-free Domain Adaptation (SFDA). We first divide the target into a trust set and a non-trust set based on prediction uncertainty. For the trust set, we directly train the target model with pseudo labels in a supervised manner. For the non-trust set, we assign a label for each sample and propose a manipulation strategy consisting of Target-guided Initialization, Semantic Feature Injection, and Domain-specific Feature Preservation, which semantically transforms the high-uncertainty sample toward the assigned category, while maintaining the generated sample in the target distribution. We progressively refined this process which simultaneously corrects pseudo-label inaccuracies in the previous trust set and decreases domain discrepancy in the previous pseudo-target domain, iteratively improving the target model. Experimental results demonstrate that our method achieves state-of-the-art performance on SFDA classification.

\section*{Acknowledgements}
This work was supported in part by the National Natural Science Foundation of China under Grant 62431017, Grant 62320106003, Grant U24A20251, Grant 62125109, Grant 62120106007, Grant 62371288, Grant 62301299, Grant 62401366, Grant 62401357, Grant 62401367, and in part by the Program of Shanghai Science and Technology Innovation Project under Grant 24BC3200800.

{
\small
\bibliographystyle{plain}
\bibliography{main}

\begin{thebibliography}{10}

\bibitem{akagic2024exploring}
Amila Akagic, Emir Buza, Medina Kapo, and Mahdi Bohlouli.
\newblock Exploring the impact of real and synthetic data in image classification: A comprehensive investigation using cifake dataset.
\newblock In {\em 2024 10th International Conference on Control, Decision and Information Technologies (CoDIT)}, pages 1207--1212. IEEE, 2024.

\bibitem{terrainc}
Sara Beery, Grant Van~Horn, and Pietro Perona.
\newblock Recognition in terra incognita.
\newblock In {\em Proceedings of the European conference on computer vision (ECCV)}, pages 456--473, 2018.

\bibitem{datum}
Yasser Benigmim, Subhankar Roy, Slim Essid, Vicky Kalogeiton, and St{\'e}phane Lathuili{\`e}re.
\newblock One-shot unsupervised domain adaptation with personalized diffusion models.
\newblock In {\em Proceedings of the 2023 IEEE/CVF Conference on Computer Vision and Pattern Recognition (CVPR)}, pages 698--708, 2023.

\bibitem{adacon}
Dian Chen, Dequan Wang, Trevor Darrell, and Sayna Ebrahimi.
\newblock Contrastive test-time adaptation.
\newblock In {\em Proceedings of the 2022 IEEE/CVF Conference on Computer Vision and Pattern Recognition (CVPR)}, pages 295--305, 2022.

\bibitem{chen2024cascade}
Yabo Chen, Jiemin Fang, Yuyang Huang, Taoran Yi, Xiaopeng Zhang, Lingxi Xie, Xinggang Wang, Wenrui Dai, Hongkai Xiong, and Qi~Tian.
\newblock {Cascade-Zero123}: One image to highly consistent {3D} with self-prompted nearby views.
\newblock In {\em Proceedings of the 18th European Conference on Computer Vision}, pages 311--330, 2024.

\bibitem{dmsfda}
Shivang Chopra, Suraj Kothawade, Houda Aynaou, and Aman Chadha.
\newblock Source-free domain adaptation with diffusion-guided source data generation.
\newblock {\em arXiv preprint arXiv:2402.04929}, 2024.

\bibitem{chu2023adversarial}
Qiaosong Chu, Shuyan Li, Guangyi Chen, Kai Li, and Xiu Li.
\newblock Adversarial alignment for source free object detection.
\newblock In {\em Proceedings of the 37th AAAI Conference on Artificial Intelligence}, pages 452--460, 2023.

\bibitem{asoge}
Chaoran Cui, Fan'an Meng, Chunyun Zhang, Ziyi Liu, Lei Zhu, Shuai Gong, and Xue Lin.
\newblock Adversarial source generation for source-free domain adaptation.
\newblock {\em IEEE Transactions on Circuits and Systems for Video Technology}, 34(6):4887--4898, 2023.

\bibitem{bnm}
Shuhao Cui, Shuhui Wang, Junbao Zhuo, Liang Li, Qingming Huang, and Qi~Tian.
\newblock Towards discriminability and diversity: Batch nuclear-norm maximization under label insufficient situations.
\newblock In {\em Proceedings of the IEEE/CVF conference on computer vision and pattern recognition}, pages 3941--3950, 2020.

\bibitem{cui2020towards}
Shuhao Cui, Shuhui Wang, Junbao Zhuo, Liang Li, Qingming Huang, and Qi~Tian.
\newblock Towards discriminability and diversity: Batch nuclear-norm maximization under label insufficient situations.
\newblock In {\em Proceedings of the IEEE/CVF conference on computer vision and pattern recognition}, pages 3941--3950, 2020.

\bibitem{cui2021fast}
Shuhao Cui, Shuhui Wang, Junbao Zhuo, Liang Li, Qingming Huang, and Qi~Tian.
\newblock Fast batch nuclear-norm maximization and minimization for robust domain adaptation.
\newblock {\em arXiv preprint arXiv:2107.06154}, 2021.

\bibitem{dhariwal2021diffusion}
Prafulla Dhariwal and Alexander Nichol.
\newblock Diffusion models beat gans on image synthesis.
\newblock In {\em Advances in Neural Information Processing Systems 34}, pages 8780--8794, 2021.

\bibitem{ding2023proxymix}
Yuhe Ding, Lijun Sheng, Jian Liang, Aihua Zheng, and Ran He.
\newblock Proxymix: Proxy-based mixup training with label refinery for source-free domain adaptation.
\newblock {\em Neural Networks}, 167:92--103, 2023.

\bibitem{ps}
Yuntao Du, Haiyang Yang, Mingcai Chen, Hongtao Luo, Juan Jiang, Yi~Xin, and Chongjun Wang.
\newblock Generation, augmentation, and alignment: A pseudo-source domain based method for source-free domain adaptation.
\newblock {\em Machine Learning}, 113(6):3611--3631, 2024.

\bibitem{everaert2024exploiting}
Martin~Nicolas Everaert, Athanasios Fitsios, Marco Bocchio, Sami Arpa, Sabine S{\"u}sstrunk, and Radhakrishna Achanta.
\newblock Exploiting the signal-leak bias in diffusion models.
\newblock In {\em Proceedings of the 2024 IEEE/CVF Winter Conference on Applications of Computer Vision (WACV)}, pages 4025--4034, 2024.

\bibitem{vlcs}
Chen Fang, Ye~Xu, and Daniel~N Rockmore.
\newblock Unbiased metric learning: On the utilization of multiple datasets and web images for softening bias.
\newblock In {\em Proceedings of the IEEE International Conference on Computer Vision}, pages 1657--1664, 2013.

\bibitem{fang2024source}
Yuqi Fang, Pew-Thian Yap, Weili Lin, Hongtu Zhu, and Mingxia Liu.
\newblock Source-free unsupervised domain adaptation: A survey.
\newblock {\em Neural Networks}, page 106230, 2024.

\bibitem{ho2020denoising}
Jonathan Ho, Ajay Jain, and Pieter Abbeel.
\newblock Denoising diffusion probabilistic models.
\newblock In {\em Advances in Neural Information Processing Systems 33}, pages 6840--6851, 2020.

\bibitem{hcl}
Jiaxing Huang, Dayan Guan, Aoran Xiao, and Shijian Lu.
\newblock Model adaptation: Historical contrastive learning for unsupervised domain adaptation without source data.
\newblock In {\em Advances in Neural Information Processing Systems 34}, pages 3635--3649, 2021.

\bibitem{imzero}
Yuyang Huang, Yabo Chen, Li~Ding, Xiaopeng Zhang, Wenrui Dai, Junni Zou, Hongkai Xiong, and Qi~Tian.
\newblock Im-zero: Instance-level motion controllable video generation in a zero-shot manner.
\newblock In {\em Proceedings of the Computer Vision and Pattern Recognition Conference}, pages 7265--7275, 2025.

\bibitem{huang2024domainfusion}
Yuyang Huang, Yabo Chen, Yuchen Liu, Xiaopeng Zhang, Wenrui Dai, Hongkai Xiong, and Qi~Tian.
\newblock {DomainFusion}: Generalizing to unseen domains with latent diffusion models.
\newblock In {\em Proceedings of the 18th European Conference on Computer Vision}, pages 480--498, 2024.

\bibitem{khachatryan2023text2video}
Levon Khachatryan, Andranik Movsisyan, Vahram Tadevosyan, Roberto Henschel, Zhangyang Wang, Shant Navasardyan, and Humphrey Shi.
\newblock {Text2Video-Zero}: Text-to-image diffusion models are zero-shot video generators.
\newblock In {\em Proceedings of the 2023 IEEE/CVF International Conference on Computer Vision (ICCV)}, pages 15954--15964, 2023.

\bibitem{koo2024flexiedit}
Gwanhyeong Koo, Sunjae Yoon, Ji~Woo Hong, and Chang~D Yoo.
\newblock Flexiedit: Frequency-aware latent refinement for enhanced non-rigid editing.
\newblock In {\em Proceedings of the 18th European Conference on Computer Vision}, pages 363--379, 2024.

\bibitem{cowa}
Jonghyun Lee, Dahuin Jung, Junho Yim, and Sungroh Yoon.
\newblock Confidence score for source-free unsupervised domain adaptation.
\newblock In {\em Proceedings of the 39th International Conference on Machine Learning}, pages 12365--12377, 2022.

\bibitem{cowajmds}
Jonghyun Lee, Dahuin Jung, Junho Yim, and Sungroh Yoon.
\newblock Confidence score for source-free unsupervised domain adaptation.
\newblock In {\em International conference on machine learning}, pages 12365--12377. PMLR, 2022.

\bibitem{lee2023few}
Suho Lee, Seungwon Seo, Jihyo Kim, Yejin Lee, and Sangheum Hwang.
\newblock Few-shot fine-tuning is all you need for source-free domain adaptation.
\newblock {\em arXiv preprint arXiv:2304.00792}, 2023.

\bibitem{li2024comprehensive}
Jingjing Li, Zhiqi Yu, Zhekai Du, Lei Zhu, and Heng~Tao Shen.
\newblock A comprehensive survey on source-free domain adaptation.
\newblock {\em IEEE Transactions on Pattern Analysis and Machine Intelligence}, 46(8):5743--5762, 2024.

\bibitem{shot}
Jian Liang, Dapeng Hu, and Jiashi Feng.
\newblock Do we really need to access the source data? source hypothesis transfer for unsupervised domain adaptation.
\newblock In {\em Proceedings of the 37th International Conference on Machine Learning}, pages 6028--6039, 2020.

\bibitem{liang2021source}
Jian Liang, Dapeng Hu, Yunbo Wang, Ran He, and Jiashi Feng.
\newblock Source data-absent unsupervised domain adaptation through hypothesis transfer and labeling transfer.
\newblock {\em IEEE Transactions on Pattern Analysis and Machine Intelligence}, 44(11):8602--8617, 2021.

\bibitem{shot++}
Jian Liang, Dapeng Hu, Yunbo Wang, Ran He, and Jiashi Feng.
\newblock Source data-absent unsupervised domain adaptation through hypothesis transfer and labeling transfer.
\newblock {\em IEEE Transactions on Pattern Analysis and Machine Intelligence}, 44(11):8602--8617, 2021.

\bibitem{bai2024zigzag}
Bai LiChen, Shitong Shao, zikai zhou, Zipeng Qi, zhiqiang xu, Haoyi Xiong, and Zeke Xie.
\newblock Zigzag diffusion sampling: Diffusion models can self-improve via self-reflection.
\newblock In {\em The Thirteenth International Conference on Learning Representations}, 2025.

\bibitem{plue}
Mattia Litrico, Alessio Del~Bue, and Pietro Morerio.
\newblock Guiding pseudo-labels with uncertainty estimation for source-free unsupervised domain adaptation.
\newblock In {\em Proceedings of the 2023 IEEE/CVF Conference on Computer Vision and Pattern Recognition (CVPR)}, pages 7640--7650, 2023.

\bibitem{litrico2023guiding}
Mattia Litrico, Alessio Del~Bue, and Pietro Morerio.
\newblock Guiding pseudo-labels with uncertainty estimation for source-free unsupervised domain adaptation.
\newblock In {\em Proceedings of the 2023 IEEE/CVF Conference on Computer Vision and Pattern Recognition (CVPR)}, pages 7640--7650, 2023.

\bibitem{liu2021source}
Yuang Liu, Wei Zhang, and Jun Wang.
\newblock Source-free domain adaptation for semantic segmentation.
\newblock In {\em Proceedings of the 2021 IEEE/CVF Conference on Computer Vision and Pattern Recognition (CVPR)}, pages 1215--1224, 2021.

\bibitem{improvedsfda}
Yu~Mitsuzumi, Akisato Kimura, and Hisashi Kashima.
\newblock Understanding and improving source-free domain adaptation from a theoretical perspective.
\newblock In {\em Proceedings of the 2024 IEEE/CVF Conference on Computer Vision and Pattern Recognition (CVPR)}, pages 28515--28524, 2024.

\bibitem{domainnet}
Xingchao Peng, Qinxun Bai, Xide Xia, Zijun Huang, Kate Saenko, and Bo~Wang.
\newblock Moment matching for multi-source domain adaptation.
\newblock In {\em Proceedings of the 2019 IEEE/CVF International Conference on Computer Vision (ICCV)}, pages 1406--1415, 2019.

\bibitem{visda}
Xingchao Peng, Ben Usman, Neela Kaushik, Judy Hoffman, Dequan Wang, and Kate Saenko.
\newblock Visda: The visual domain adaptation challenge.
\newblock {\em arXiv preprint arXiv:1710.06924}, 2017.

\bibitem{sdxl}
Dustin Podell, Zion English, Kyle Lacey, Andreas Blattmann, Tim Dockhorn, Jonas M{\"u}ller, Joe Penna, and Robin Rombach.
\newblock {SDXL}: Improving latent diffusion models for high-resolution image synthesis.
\newblock In {\em The Twelfth International Conference on Learning Representations}, 2024.

\bibitem{qiu2023freenoise}
Haonan Qiu, Menghan Xia, Yong Zhang, Yingqing He, Xintao Wang, Ying Shan, and Ziwei Liu.
\newblock Freenoise: Tuning-free longer video diffusion via noise rescheduling.
\newblock In {\em The Twelfth International Conference on Learning Representations}, 2024.

\bibitem{cpga}
Zhen Qiu, Yifan Zhang, Hongbin Lin, Shuaicheng Niu, Yanxia Liu, Qing Du, and Mingkui Tan.
\newblock Source-free domain adaptation via avatar prototype generation and adaptation.
\newblock In {\em Proceedings of the 30th International Joint Conference on Artificial Intelligence (IJCAI-21)}, pages 2921--2927, 2021.

\bibitem{rombach2022high}
Robin Rombach, Andreas Blattmann, Dominik Lorenz, Patrick Esser, and Bj{\"o}rn Ommer.
\newblock High-resolution image synthesis with latent diffusion models.
\newblock In {\em Proceedings of the 2022 IEEE/CVF Conference on Computer Vision and Pattern Recognition (CVPR)}, pages 10684--10695, 2022.

\bibitem{dreambooth}
Nataniel Ruiz, Yuanzhen Li, Varun Jampani, Yael Pritch, Michael Rubinstein, and Kfir Aberman.
\newblock Dreambooth: Fine tuning text-to-image diffusion models for subject-driven generation.
\newblock In {\em Proceedings of the IEEE/CVF conference on computer vision and pattern recognition}, pages 22500--22510, 2023.

\bibitem{office31}
Kate Saenko, Brian Kulis, Mario Fritz, and Trevor Darrell.
\newblock Adapting visual category models to new domains.
\newblock In {\em Proceedings of the 11th European Conference on Computer Vision}, pages 213--226, 2010.

\bibitem{song2020denoising}
Jiaming Song, Chenlin Meng, and Stefano Ermon.
\newblock Denoising diffusion implicit models.
\newblock In {\em The Ninth International Conference on Learning Representations}, 2021.

\bibitem{tpds}
Song Tang, An~Chang, Fabian Zhang, Xiatian Zhu, Mao Ye, and Changshui Zhang.
\newblock Source-free domain adaptation via target prediction distribution searching.
\newblock {\em International Journal of Computer Vision}, 132(3):654--672, 2024.

\bibitem{gkd}
Song Tang, Yuji Shi, Zhiyuan Ma, Jian Li, Jianzhi Lyu, Qingdu Li, and Jianwei Zhang.
\newblock Model adaptation through hypothesis transfer with gradual knowledge distillation.
\newblock In {\em 2021 IEEE/RSJ International Conference on Intelligent Robots and Systems (IROS)}, pages 5679--5685, 2021.

\bibitem{prode}
Song Tang, Wenxin Su, Yan Gan, Mao Ye, Jianwei~Dr. Zhang, and Xiatian Zhu.
\newblock Proxy denoising for source-free domain adaptation.
\newblock In {\em The Thirteenth International Conference on Learning Representations}, 2025.

\bibitem{difo}
Song Tang, Wenxin Su, Mao Ye, and Xiatian Zhu.
\newblock Source-free domain adaptation with frozen multimodal foundation model.
\newblock In {\em Proceedings of the 2024 IEEE/CVF Conference on Computer Vision and Pattern Recognition (CVPR)}, pages 23711--23720, 2024.

\bibitem{sclm}
Song Tang, Yan Zou, Zihao Song, Jianzhi Lyu, Lijuan Chen, Mao Ye, Shouming Zhong, and Jianwei Zhang.
\newblock Semantic consistency learning on manifold for source data-free unsupervised domain adaptation.
\newblock {\em Neural Networks}, 152:467--478, 2022.

\bibitem{thomas2025s}
Xavier Thomas and Deepti Ghadiyaram.
\newblock What's in a latent? leveraging diffusion latent space for domain generalization.
\newblock {\em arXiv preprint arXiv:2503.06698}, 2025.

\bibitem{officehome}
Hemanth Venkateswara, Jose Eusebio, Shayok Chakraborty, and Sethuraman Panchanathan.
\newblock Deep hashing network for unsupervised domain adaptation.
\newblock In {\em Proceedings of the 2017 IEEE Conference on Computer Vision and Pattern Recognition}, pages 5018--5027, 2017.

\bibitem{wang2018deep}
Mei Wang and Weihong Deng.
\newblock Deep visual domain adaptation: A survey.
\newblock {\em Neurocomputing}, 312:135--153, 2018.

\bibitem{wang2021source}
Yuxi Wang, Jian Liang, and Zhaoxiang Zhang.
\newblock Source data-free cross-domain semantic segmentation: Align, teach and propagate.
\newblock {\em arXiv e-prints}, pages arXiv--2106, 2021.

\bibitem{wilson2020survey}
Garrett Wilson and Diane~J Cook.
\newblock A survey of unsupervised deep domain adaptation.
\newblock {\em ACM Transactions on Intelligent Systems and Technology (TIST)}, 11(5):1--46, 2020.

\bibitem{wu2023freeinit}
Tianxing Wu, Chenyang Si, Yuming Jiang, Ziqi Huang, and Ziwei Liu.
\newblock {FreeInit}: Bridging initialization gap in video diffusion models.
\newblock In {\em Proceedings of the 18th European Conference on Computer Vision}, pages 378--394, 2024.

\bibitem{xie2024sanaefficienthighresolutionimage}
Enze Xie, Junsong Chen, Junyu Chen, Han Cai, Haotian Tang, Yujun Lin, Zhekai Zhang, Muyang Li, Ligeng Zhu, Yao Lu, and Song Han.
\newblock {SANA}: Efficient high-resolution image synthesis with linear diffusion transformers.
\newblock In {\em The Thirteenth International Conference on Learning Representations}, 2025.

\bibitem{xu2025stylessp}
Ruojun Xu, Weijie Xi, XiaoDi Wang, Yongbo Mao, and Zach Cheng.
\newblock Stylessp: Sampling startpoint enhancement for training-free diffusion-based method for style transfer.
\newblock In {\em Proceedings of the Computer Vision and Pattern Recognition Conference}, pages 18260--18269, 2025.

\bibitem{yang2022divide}
Jianfei Yang, Xiangyu Peng, Kai Wang, Zheng Zhu, Jiashi Feng, Lihua Xie, and Yang You.
\newblock Divide to adapt: Mitigating confirmation bias for domain adaptation of black-box predictors.
\newblock In {\em The Eleventh International Conference on Learning Representations}, 2023.

\bibitem{ada}
Shiqi Yang, Shangling Jui, Joost Van De~Weijer, et~al.
\newblock Attracting and dispersing: A simple approach for source-free domain adaptation.
\newblock In {\em Advances in Neural Information Processing Systems 35}, pages 5802--5815, 2022.

\bibitem{nrc}
Shiqi Yang, Joost Van~de Weijer, Luis Herranz, Shangling Jui, et~al.
\newblock Exploiting the intrinsic neighborhood structure for source-free domain adaptation.
\newblock In {\em Advances in Neural Information Processing Systems 34}, pages 29393--29405, 2021.

\bibitem{yang2023casting}
Shiqi Yang, Yaxing Wang, Luis Herranz, Shangling Jui, and Joost van~de Weijer.
\newblock Casting a bait for offline and online source-free domain adaptation.
\newblock {\em Computer Vision and Image Understanding}, 234:103747, 2023.

\bibitem{gsfda}
Shiqi Yang, Yaxing Wang, Joost Van De~Weijer, Luis Herranz, and Shangling Jui.
\newblock Generalized source-free domain adaptation.
\newblock In {\em Proceedings of the IEEE/CVF international conference on computer vision}, pages 8978--8987, 2021.

\bibitem{ye2021source}
Mucong Ye, Jing Zhang, Jinpeng Ouyang, and Ding Yuan.
\newblock Source data-free unsupervised domain adaptation for semantic segmentation.
\newblock In {\em Proceedings of the 29th ACM International Conference on Multimedia}, pages 2233--2242, 2021.

\bibitem{elr}
Li~Yi, Gezheng Xu, Pengcheng Xu, Jiaqi Li, Ruizhi Pu, Charles Ling, Ian McLeod, and Boyu Wang.
\newblock When source-free domain adaptation meets learning with noisy labels.
\newblock In {\em The Eleventh International Conference on Learning Representations}, 2023.

\bibitem{zhang2023tale}
Junyi Zhang, Charles Herrmann, Junhwa Hur, Luisa~Polania Cabrera, Varun Jampani, Deqing Sun, and Ming-Hsuan Yang.
\newblock A tale of two features: Stable diffusion complements {DINO} for zero-shot semantic correspondence.
\newblock In {\em Advances in Neural Information Processing Systems 36}, pages 45533--45547, 2023.

\bibitem{zhang2023controlvideo}
Yabo Zhang, Yuxiang Wei, Dongsheng Jiang, Xiaopeng Zhang, Wangmeng Zuo, and Qi~Tian.
\newblock {ControlVideo}: Training-free controllable text-to-video generation.
\newblock In {\em The Twelfth International Conference on Learning Representations}, 2024.

\bibitem{zhang2024videoelevator}
Yabo Zhang, Yuxiang Wei, Xianhui Lin, Zheng Hui, Peiran Ren, Xuansong Xie, and Wangmeng Zuo.
\newblock Videoelevator: Elevating video generation quality with versatile text-to-image diffusion models.
\newblock In {\em Proceedings of the AAAI Conference on Artificial Intelligence}, volume~39, pages 10266--10274, 2025.

\bibitem{crs}
Yixin Zhang, Zilei Wang, and Weinan He.
\newblock Class relationship embedded learning for source-free unsupervised domain adaptation.
\newblock In {\em Proceedings of the 2023 IEEE/CVF Conference on Computer Vision and Pattern Recognition (CVPR)}, pages 7619--7629, 2023.

\bibitem{zhang2022divide}
Ziyi Zhang, Weikai Chen, Hui Cheng, Zhen Li, Siyuan Li, Liang Lin, and Guanbin Li.
\newblock Divide and contrast: Source-free domain adaptation via adaptive contrastive learning.
\newblock In {\em Advances in Neural Information Processing Systems 35}, pages 5137--5149, 2022.

\bibitem{zhao2023unleashing}
Wenliang Zhao, Yongming Rao, Zuyan Liu, Benlin Liu, Jie Zhou, and Jiwen Lu.
\newblock Unleashing text-to-image diffusion models for visual perception.
\newblock In {\em Proceedings of the 2023 IEEE/CVF International Conference on Computer Vision (ICCV)}, pages 5706--5716, 2023.

\bibitem{zheng2024cogview3}
Wendi Zheng, Jiayan Teng, Zhuoyi Yang, Weihan Wang, Jidong Chen, Xiaotao Gu, Yuxiao Dong, Ming Ding, and Jie Tang.
\newblock {CogView3}: Finer and faster text-to-image generation via relay diffusion.
\newblock In {\em Proceedings of the 18th European Conference on Computer Vision}, pages 1--22, 2024.

\bibitem{cpd}
Lihua Zhou, Nianxin Li, Mao Ye, Xiatian Zhu, and Song Tang.
\newblock Source-free domain adaptation with class prototype discovery.
\newblock {\em Pattern Recognition}, 145:109974, 2024.

\end{thebibliography}
}

\newpage
\begin{center}
\centering
{\Large \textbf{Appendix}}
\end{center}

\appendix

\renewcommand{\thefigure}{A.\arabic{figure}} 
\renewcommand{\thetable}{A.\arabic{table}}   
\setcounter{figure}{0} 
\setcounter{table}{0}  

\section{Evaluation Datasets}

We evaluate our method on four Source-free Domain Adaptation (SFDA) datasets benchmarks, including the small-scale Office-31 dataset~\cite{office31}, the medium-scale Office-Home dataset~\cite{officehome}, and two large-scale datasets (\emph{i.e.}, VisDA~\cite{visda} and DomainNet-126~\cite{domainnet}). The complete dataset statistics and domain configurations are elaborated on below.

\textbf{Office-31~\cite{office31}:} This small-scale dataset contains 31 object categories commonly found in office environments (e.g., keyboards, laptops, file cabinets). It comprises 4,652 images across 3 domains: \textit{Amazon}, \textit{Webcam}, and \textit{DSLR}.

\textbf{Office-Home~\cite{officehome}:} Office-Home is a more challenging medium-scale domain adaptation dataset with 65 categories and 15,500 images distributed across 4 domains: \textit{Art}, \textit{Clipart}, \textit{Product}, and \textit{Real-World}.

\textbf{VisDA~\cite{visda}:} VisDA is a large-scale dataset for domain adaptation, originally designed for the 2017 Visual Domain Adaptation Challenge. It focuses on synthetic-to-real transfer with 12 object categories. The source domain contains 152,397 synthetic images, while the target domain has 55,388 real-world images.

\textbf{DomainNet-126~\cite{domainnet}:} DomainNet-126 is a subset of the full DomainNet dataset~\cite{domainnet}, curated for domain adaptation research. It includes 126 object categories across 4 domains:  \textit{Clipart}, \textit{Painting}, \textit{Real}, and \textit{Sketch}. The dataset contains 145k images, making it one of the largest SFDA benchmarks.

\section{Implementation Details}

\subsection{Source Model Pre-training}
\label{sourcetrain}
For the source model training, we employ only the cross-entropy loss as the objective function. Regarding the training hyperparameters, we set weight\_decay (weight decay, L2 regularization coefficient) to 5e-4, lr\_gamma (learning rate gamma, learning rate scheduler parameter) to 0.0003, lr\_decay (learning rate decay rate) to 0.75, and momentum (SGD momentum parameter) to 0.9. For the large-scale DomainNet-126 and VisDA, to facilitate the model convergence, we set batch size as 128, n\_iter\_per\_epoch (Number of iterations per epoch ) as 200, n\_epoch (Total training epochs) as 100, lr (learning rate) as 3e-3. For other datasets, we set batch size as 32, n\_iter\_per\_epoch as 300, n\_epoch as 50, and lr as 1e-3. As for source model architecture, we employ ResNet-50 for Office-31~\cite{office31}, Office-Home~\cite{officehome} and 
DomainNet-126~\cite{domainnet}, and ResNet-101 for VisDA~\cite{visda}. 

\subsection{DPTM Details}
 
\textbf{Method-related Hyperparameters and Configurations.}
\label{methodhyperparameters}
 We employ stable-diffusion v1-5~\cite{rombach2022high} as the diffusion model to generate 512$\times$512 images with 20 denoising steps. $\gamma_1=5.5$ and $\gamma_2=0$. We set the threshold $E$ to 0.01, and the total refinement iteration count $R$ to 10. Note that setting $E$ and $R$ to other values may obtain superior performance. 

\textbf{Training-related Hyperparameters and Configurations.} For the target model architecture, we adopt ResNet-101 for VisDA~\cite{visda} and also for the other datasets. To further enhance the performance of the target model while mitigating the inevitable domain discrepancy between pseudo-target and real target domains, we incorporate the baseline Unsupervised Domain Adaptation (UDA) method BNM~\cite{bnm} during the target model training with pseudo-target data. This ensures that even when the samples generated by our method are not fully aligned with the real target distribution, the inconsistency can be further alleviated through the BNM adaptation process. For training hyperparameters, all parameters, including weight\_decay, lr\_gamma, lr\_decay, momentum, n\_iter\_per\_epoch, and n\_epoch as 100, and lr remains unchanged as Source Model Pre-training in Section~\ref{sourcetrain}. 

\textbf{Experiments Compute Resources.}
For Office-31~\cite{office31} and Office-Home~\cite{officehome}, all related experiments are conducted using a single NVIDIA Tesla V100. For the large-scale Domain VisDA~\cite{visda} and DomainNet-126~\cite{domainnet}, all related experiments are conducted using a single NVIDIA Tesla H100.

\section{Supplementary Results}
\subsection{Full Results on VisDA}
We present full results on VisDA in Table~\ref{tab:full_vis}. We report accuracy results over 12 categories and report the per-class accuracy. Notably, we also reproduced the results using only the source model. Experimental results demonstrate that our method outperforms existing SOTA methods across all categories. Moreover, our method demonstrates two significant advantages: (1) For categories such as plane, horse, knife, plant, and skateboard, our method achieves near-perfect classification accuracy, exceeding 99\%. (2) For challenging categories like truck and car, where existing methods perform poorly, our approach substantially improves accuracy to over 90\%. For the truck category, where most methods exhibit poor performance, our method exceeds the current SOTA method by 24.1\%.

\begin{table}[!t]
\renewcommand\tabcolsep{3.6pt}
\renewcommand\arraystretch{1.0}
\centering
\caption{Full Results~(\%) on \textbf{VisDA} evaluated with ResNet-101. The top three performances in each column are highlighted in red, orange, and yellow, respectively.}\label{tab:full_vis}
\scriptsize
\begin{tabular}{@{} l l |  c c c c c c c c c c c c c @{}}
        \toprule
        \multirow{2}{*}{Method} &\multirow{2}{*}{Venue} 
        &\multicolumn{13}{c}{\textbf{VisDA}} \\
        & &plane &bcycl &bus &car &horse &knife &mcycl &person &plant &sktbrd &train &truck &Perclass \\
        \midrule
        Source        &--  
        &92.3 &33.3 &76.4 &60.9  &86.5 &32.7 &89.9 &33.3 &79.8 &48.0 &87.7 &18.4 &61.6  \\
        \midrule
        CPGA~\cite{cpga}        &IJCAI21  
        &95.6 &89.0 &75.4 &64.9 &91.7 &\besttwo{97.5} &89.7 &83.8 &93.9 &93.4 &87.7 &\besttwo{69.0} &86.0 \\
        ASOGE~\cite{asoge}       &TCSVT23  
        &94.9	&84.3	&76.8	&54.3	&94.9	&93.4	&86.0	&85.0	&87.2	&90.0	&86.7	&62.7 &83.2 
        \\
        ISFDA~\cite{improvedsfda}        &CVPR24  
        &97.5 &91.4 &87.9 &79.4 &97.2 &97.2 &\besttwo{92.2} &83.0 &\bestthree{96.4} &94.2 &91.1 &53.0 &88.4  \\
        PS~\cite{ps}        &ML24  
        &95.3 &86.2 &82.3 &61.6 &93.3 &95.7 &86.7 &80.4 &91.6 &90.9 &86.0 &59.5 &84.1 \\
        DM-SFDA~\cite{dmsfda} &-- &\besttwo{98.1} &89.8 &\besttwo{90.6} &\besttwo{90.5} &96.8 &95.2 &\besttwo{92.2} &\besttwo{93.4} &\besttwo{97.8} &\bestthree{94.4} &92.4 &48.8 &86.3\\
        \midrule
        SHOT~\cite{shot}   &ICML20 &95.0 &87.4 &80.9 &57.6 &93.9 &94.1 &79.4 &80.4 &90.9 &89.8 &85.8 &57.5 &82.7 \\
        NRC~\cite{nrc}      &NIPS21    &96.8 &91.3 &82.4 &62.4 &96.2 &95.9 &86.1 &\bestthree{90.7} &94.8 &94.1 &90.4 &59.7 &85.9 \\
        GKD~\cite{gkd}      &IROS21   &95.3 &87.6 &81.7 &58.1 &93.9 &94.0 &80.0 &80.0 &91.2 &91.0 &86.9 &56.1 &83.0 \\ 
        AaD~\cite{ada}      &NIPS22    &97.4 &90.5 &80.8 &76.2 &\bestthree{97.3} &96.1 &89.8 &82.9 &95.5 &93.0 &92.0 &64.7 &88.0 \\ 
        AdaCon~\cite{adacon}   &CVPR22    &97.0 &84.7 &84.0 &77.3 &96.7 &93.8 &\bestthree{91.9} &84.8 &94.3 &93.1 &\besttwo{94.1} &49.7 &86.8 \\ 
        CoWA~\cite{cowa}     &ICML22    &96.2 &89.7 &83.9 &73.8 &96.4 &\bestthree{97.4} &89.3 &86.8 &94.6 &92.1 &88.7 &53.8 &86.9 \\
        SCLM~\cite{sclm}     &NN22    
        &97.1 &90.7 &85.6 &62.0 &\bestthree{97.3} &94.6 &81.8 &84.3 &93.6 &92.8 &88.0 &55.9 &85.3 \\
        ELR~\cite{elr}      &ICLR23    &97.1 &89.7 &82.7 &62.0 &96.2 &97.0 &87.6 &81.2 &93.7 &94.1 &90.2 &58.6 &85.8 \\ 
        PLUE~\cite{plue}     &CVPR23    &94.4 &\besttwo{91.7} &89.0 &70.5 &96.6 &94.9 &\besttwo{92.2} &88.8 &92.9 &\besttwo{95.3} &91.4 &61.6 &88.3 \\ 
        CPD~\cite{cpd}      &PR24      &96.7 &88.5 &79.6 &69.0 &95.9 &96.3 &87.3 &83.3 &94.4 &92.9 &87.0 &58.7 &85.5 \\ 
        TPDS~\cite{tpds}     &IJCV24    &\bestthree{97.6} &\bestthree{91.5} &\bestthree{89.7} &83.4 &\besttwo{97.5} &96.3 &\besttwo{92.2} &82.4 &96.0 &94.1 &90.9 &40.4 &87.6 \\ 
        DIFO~\cite{difo}  &CVPR24     &\bestthree{97.6} &88.7 &83.7 &\bestthree{80.8} &95.9 &95.3 &\bestthree{91.9} &85.0 &89.4 &93.2 &\bestthree{93.2} &\besttwo{69.0} &\bestthree{88.6} \\
        ProDe~\cite{prode}   &ICLR25    &96.6 &90.3 &83.9 &80.2 &96.1 &96.9 &90.3 &86.4 &90.8 &94.0 &91.3 &\bestthree{67.0} &\besttwo{88.7} \\
        \midrule
         \textbf{DPTM (ours)}   &--   &\best{99.5}  &\best{97.1} &\best{96.2} &\best{93.0} &\best{99.2}  &\best{99.2} &\best{98.8} &\best{97.7} &\best{99.3} &\best{99.7}  &\best{98.6} &\best{93.1} &\best{97.6} 
     \\
    \bottomrule
    \end{tabular}
\end{table}

\textbf{Manipulation Mechanism of Non-trust Set.} 
We present ablation studies on different components of the Manipulation Mechanism of Non-trust Set via visualization results. As is demonstrated in Table 4, SDXL and SD15 present comparable results. To present better visualization, we employ SDXL and set the denoising steps to 50. We present images generated by: (a) $\mathbf{x}_l^u$ (b) $\tilde{\mathbf{x}}_l^{u}$ w/o Target-guided Initialization (c) $\tilde{\mathbf{x}}_l^{u}$ w/o Semantic Feature Injection (d) $\tilde{\mathbf{x}}_l^{u}$ w/o Domain-specific Feature Preservation (e) $\tilde{\mathbf{x}}_l^{u}$ of our method, respectively. As shown in Figure 2: (1) Our method's manipulated samples $\tilde{\mathbf{x}}_l^{u}$, exhibit the best semantic alignment with their assigned labels $\hat{y}_l=$ and the best preservation of target distribution characteristics. (2) Column (b) (d) (e) that involve Semantic Feature Injection transforms the original semantics to the assigned label successfully, while Column (c) w/o Semantic Feature Injection achieves poor alignment with the assigned label, demonstrating the effectiveness of Semantic Feature Injection.
(3) Column (b) excludes Target-guided Initialization, and only Domain-specific Feature Preservation works for maintaining the images within the target distribution. As a result, Column (b) preserves target domain features worse than Column (e). Similarly, Column (d) excludes Domain-specific Feature Preservation, and only Target-guided Initialization works, also exhibiting worse preservation of target domain features than Column (e). These results demonstrate the effectiveness of Target-guided Initialization and Domain-specific Feature Preservation.

\textbf{The SFDA Model Size.} Table~\ref{tab:full_vis} shows results on VisDA with ResNet-101. To further demonstrate the superior performance of our method, we provide extra results with ResNet-50 in Table~\ref{tab:rn50}. These results demonstrate that: (1) Our method exhibits robustness to model size. It maintains high performance even when using a smaller ResNet-50 target SFDA model. Notably, our method with ResNet-50 even outperforms existing comparative methods that use a larger ResNet-101 backbone, highlighting its superior
adaptation performance regardless of model scale.
Our method is also scalable with respect to the target SFDA model size. When using a larger ResNet-101 target SFDA model, our method achieves
better performance compared to using ResNet-50, suggesting that its effectiveness can scale with increased target SFDA model size.

\begin{table}[!t]
\renewcommand\tabcolsep{3.6pt}
\renewcommand\arraystretch{1.0}
\centering
\caption{Full Results~(\%) on \textbf{VisDA} evaluated with ResNet-50. }\label{tab:rn50}
\footnotesize 
\begin{tabular}{@{} l |  c c c c c c c c c c c c c @{}}
        \toprule
        Method 
        &plane &bcycl &bus &car &horse &knife &mcycl &person &plant &sktbrd &train &truck &Perclass \\
        \midrule
Source-R50 & 79.7  & 35.5  & 44.5 & 63.8 & 62.0  & 25.0  & 86.9  & 26.6   & 77.6  & 30.0   & 94.7  & 12.7  & 53.3     \\
Ours-R50   & 99.5  & 96.1  & 93.5 & 82.5 & 98.3  & 99.2  & 96.3  & 96.8   & 98.8  & 98.5   & 98.1  & 81    & 94.9  \\
\bottomrule
\end{tabular}
\end{table}

\begin{table*}[!t]
\renewcommand{\baselinestretch}{1.0}
\renewcommand\arraystretch{1.0}
\setlength{\tabcolsep}{1pt}
\setlength{\abovecaptionskip}{0pt}
\centering
\caption{Comparison results with DATUM on Office-Home dataset.}
\label{datum}
\resizebox{\textwidth}{!}{
\footnotesize
\begin{tabular}{@{}l|ccccccccccccc@{}}
\toprule
Method & Ar$\to$Cl & Ar$\to$Pr & Ar$\to$Rw
& Cl$\to$Ar & Cl$\to$Pr & Cl$\to$Rw & Pr$\to$Ar & Pr$\to$Cl & Pr$\to$Rw  
& Rw$\to$Ar & Rw$\to$Cl & Rw$\to$Pr & Avg. \\
\midrule
Source      & 50.1 & 67.9 & 74.4 & 55.2 & 65.2 & 67.2 & 53.4 & 44.5 & 74.1 & 64.2 & 51.5 & 78.7 & 62.2 \\
DATUM       & 55.3 & 76.8 & 79.3 & 65.1 & 77.7 & 78.6 & 62.4 & 52.1 & 79.7 & 66.6 & 55.9 & 80.5 & 69.2 \\
\midrule
\textbf{DPTM (ours)} & 86.7 & 94.2 & 92.8 & 91.5 & 94.0 & 92.6 & 90.6 & 86.4 & 92.8 & 90.5 & 87.1 & 94.7 & 91.2 \\
\bottomrule
\end{tabular}
}
\end{table*}

\textbf{Different Components of DPTM.} We  conduct ablation studies on the independent usage of each individual component. For clarity in the tables, we refer to Target-guided Initialization, Semantic Feature Injection, and Domain-specific Feature Preservation as TGI, SFI, and DFP, respectively. We present comprehensive component-wise ablation results, including the performance of the model with only one component enabled and with each component individually removed. The results are shown in Table~\ref{componentwise}, further demonstrating the effectiveness of our method. 

\begin{table}[!t]
\renewcommand{\baselinestretch}{1.0}
\renewcommand\arraystretch{1.0}
\renewcommand\tabcolsep{1pt}
\setlength{\abovecaptionskip}{0pt}
\centering
\caption{Component-wise ablation studies on Office-Home dataset.}
\label{componentwise}
\footnotesize 
\resizebox{\textwidth}{!}{
\begin{tabular}{@{} ccc |  c c c c c c c c c c c c c@{}}
\toprule
TGI & SFI & DFP & Ar$\to$Cl & Ar$\to$Pr & Ar$\to$Rw & Cl$\to$Ar & Cl$\to$Pr & Cl$\to$Rw & Pr$\to$Ar & Pr$\to$Cl & Pr$\to$Rw & Rw$\to$Ar & Rw$\to$Cl & Rw$\to$Pr & Avg. \\
\midrule
\XSolidBrush   & \XSolidBrush   & \XSolidBrush   & 50.1  & 67.9  & 74.4  & 55.2  & 65.2  & 67.2  & 53.4  & 44.5  & 74.1  & 64.2  & 51.5  & 78.7  & 62.2 \\
\Checkmark   & \XSolidBrush   & \XSolidBrush   & 69.2  & 86.2  & 82.2  & 74.6  & 87.8  & 80.7  & 76.7  & 67.8  & 82.4  & 73.5  & 66.9  & 87.1  & 77.9 \\
\XSolidBrush   & \Checkmark   & \XSolidBrush   & 59.6  & 80.8  & 82.3  & 67.9  & 83.1  & 80.2  & 64.6  & 68.1  & 81.7  & 70.6  & 67.9  & 86.5  & 74.4 \\
\XSolidBrush   & \XSolidBrush   & \Checkmark   & 67.8  & 87.6  & 82.7  & 68.7  & 84.0  & 80.6  & 70.7  & 67.6  & 82.4  & 74.1  & 67.1  & 88.7  & 76.8 \\
\XSolidBrush   & \Checkmark   & \Checkmark   & 75.9  & 89.4  & 88.3  & 78.9  & 87.9  & 87.0  & 76.0  & 74.1  & 87.8  & 80.0  & 75.9  & 89.1  & 82.5 \\
\Checkmark   & \XSolidBrush   & \Checkmark   & 72.9  & 88.0  & 84.1  & 77.5  & 88.3  & 83.6  & 76.8  & 69.0  & 84.0  & 75.7  & 68.8  & 88.4  & 79.8 \\
\Checkmark   & \Checkmark   & \XSolidBrush   & 70.2  & 89.9  & 85.5  & 81.1  & 89.4  & 89.3  & 80.7  & 70.7  & 86.3  & 80.3  & 72.1  & 91.0  & 82.2 \\
\Checkmark   & \Checkmark   & \Checkmark   & 86.7  & 94.2  & 92.8  & 91.5  & 94.0  & 92.6  & 90.6  & 86.4  & 92.8  & 90.5  & 87.1  & 94.7  & 91.2 \\
    \bottomrule
    \end{tabular}
}
\end{table}

\section{Supplementary Analyses}

\subsection{Analysis on Trust and Non-trust Partition for Target Domain.} 
\textbf{Reliability of entropy-based trust and non-trust partition.}
Entropy-based selection of reliable pseudo-labels of target samples is commonly used in SFDA, and its effectiveness is demonstrated in prior studies~\cite{liang2021source}. To further evaluate its reliability, we report the trust set accuracy evolving with $r$ on the Office-Home dataset as Table~\ref{fullacc} ($r$ denotes the $r$-th refinement iteration, where $r=1,2,...R$, and in our experiments we set $R=10$). Figure 2(a) shows that the size of the trust set grows as $r$ increases. Remarkably, the trust accuracy remains consistently high across all tasks using a total of 10 refinement iterations. Besides, when $r$ is small, the trust set accuracy may not be high, but will increase to a high value as $r$ grows. For example, in tasks like Pr$\to$Ar and Pr$\to$Cl, trust accuracy is below 90\% when $r=2$ but reaches higher than 97\% when $r=10$. These results validate the effectiveness of our method, which could correct errors in the trust set with the growth of $r$. Note that the non-trust set accuracy does not affect the performance of our method, as we completely discard the original pseudo-labels of non-trust samples.

\begin{table}[!t]
\renewcommand{\baselinestretch}{1.0}
\renewcommand\arraystretch{1.0}
\renewcommand\tabcolsep{1pt}
\setlength{\abovecaptionskip}{0pt}
\centering
\caption{The trust set accuracy evolving with $r$ on the Office-Home dataset.}
\label{fullacc}
\footnotesize 
\begin{tabular}{@{}l  |  c c c c c c c c c c c c @{}}
\toprule
$r$&Ar$\to$Cl &Ar$\to$Pr &Ar$\to$Rw
&Cl$\to$Ar &Cl$\to$Pr &Cl$\to$Rw  &Pr$\to$Ar &Pr$\to$Cl &Pr$\to$Rw  
&Rw$\to$Ar &Rw$\to$Cl &Rw$\to$Pr  \\
\midrule
$1$  & 95.4  & 97.6  & 98.5  & 99.1  & 100.0 & 99.8  & 100.0 & 100.0 & 99.8  & 100.0 & 97.4  & 99.1  \\
$2$  & 85.3  & 95.0  & 97.2  & 88.8  & 93.9  & 96.1  & 85.9  & 83.8  & 96.6  & 87.5  & 86.4  & 96.3  \\
$3$  & 94.5  & 98.3  & 97.6  & 91.2  & 98.2  & 97.3  & 90.0  & 91.9  & 97.8  & 92.9  & 93.1  & 98.2  \\
$4$  & 94.2  & 98.0  & 98.1  & 94.3  & 98.1  & 97.8  & 93.6  & 95.2  & 98.0  & 94.4  & 95.7  & 98.4  \\
$5$  & 96.6  & 99.0  & 98.7  & 95.1  & 99.0  & 98.5  & 94.4  & 95.8  & 98.2  & 95.6  & 95.7  & 99.0    \\
$6$  & 96.3  & 99.1  & 98.6  & 95.7  & 99.1  & 98.6  & 96.1  & 96.8  & 98.6  & 95.0  & 97.3  & 99.0    \\
$7$  & 97.1  & 99.4  & 98.5  & 96.9  & 99.4  & 98.5  & 97.1  & 96.8  & 98.7  & 96.8  & 96.6  & 99.2  \\
$8$  & 97.2  & 99.3  & 99.2  & 97.0  & 99.2  & 99.2  & 96.6  & 97.9  & 98.8  & 95.7  & 97.2  & 99.5  \\
$9$  & 98.6  & 99.5  & 99.2  & 97.0  & 99.6  & 99.2  & 97.0  & 98.1  & 98.9  & 97.3  & 97.8  & 99.3  \\
$10$ & 97.5  & 99.4  & 98.7  & 97.5  & 99.6  & 99.1  & 97.2  & 97.7  & 99.4  & 97.3  & 98.5  & 99.5 \\
    \bottomrule
    \end{tabular}
\end{table}

\textbf{Set all target samples to the non-trust set.}
The high performance gain of our method stems from two key mechanisms: i) progressively expanding the trust set with high-accuracy pseudo-labels to allow the SFDA model to learn real target domain features, and ii) progressively reducing the manipulated non-trust set. Mechanism ii) is critical to prevent features from the synthetic domain from becoming dominant, since there is an inherent gap that persists between the synthetic and real target domains (even though we employ alignments to bridge the gap). This phenomenon fundamentally stems from the inherent domain gap between synthetic and real data, a well-documented challenge that has been rigorously demonstrated in prior work~\cite{akagic2024exploring}. Therefore, canceling the trust set could cause degraded performance, since we could only obtain features from the synthetic domain. We validate it with an empirical study on the first 4 Office-Home tasks (Ar$\to$Cl, Ar$\to$Pr, Ar$\to$Rw, and Cl$\to$Ar), where we cancel the trust set and set all target samples to the non-trust set.  $r$ denotes the $r$-th refinement iteration, and we report results of $r=1,2,...,5$. As shown in Table~\ref {allnon}, when using only non-trust samples, the performance is not improved as $r$ grows. 

\begin{table}[!t]
\renewcommand{\baselinestretch}{1.0}
\renewcommand\arraystretch{1.0}
\renewcommand\tabcolsep{1pt}
\setlength{\abovecaptionskip}{0pt}
\centering
\caption{The performance trajectory of setting all target samples to the non-trust set on the Office-Home dataset.}
\label{allnon}
\footnotesize 
\begin{tabular}{@{}l  |  c c c c @{}}
\toprule
$r$&Ar$\to$Cl &Ar$\to$Pr &Ar$\to$Rw
&Cl$\to$Ar  \\
\midrule
0   & 50.1      & 67.9      & 74.4      & 55.2      \\
1   & 56.4      & 78.7      & 81.7      & 63.0      \\
2   & 55.9      & 78.5      & 81.4      & 62.8      \\
3   & 55.9      & 78.3      & 81.6      & 62.7      \\
4   & 55.8      & 78.1      & 81.4      & 62.7      \\
5   & 55.7      & 77.9      & 81.3      & 62.7      \\
    \bottomrule
    \end{tabular}
\end{table}

\subsection{Analysis on Manipulation of Non-trust Set.} 
\textbf{Random assignment of labels for non-trust samples.} From a performance perspective, we employ random label assignment, instead of using the original pseudo-labels of non-trust samples given by the SFDA model, for two key reasons. First, there are no obvious patterns between the original pseudo-labels and ground truth labels of non-trust samples, and the original pseudo-labels cannot provide valid semantic priors. Second, if we use the diffusion model to semantically transform non-trust samples to their original pseudo-labels, this approach could inevitably introduce the class imbalance issue in the manipulated non-trust set. Therefore, we adopt a uniform random label reassignment strategy as formalized in Equation (2). To validate this claim, we perform comparative experiments using the original pseudo-labels for generation. Due to time limits, the experiments were performed on the first 4 Office-Home tasks (Ar$\to$Cl, Ar$\to$Pr, Ar$\to$Rw, and Cl$\to$Ar). The results presented in Table~\ref{random} demonstrate the validity of our choice. Besides, from the perspective of reproducibility and stability, the results are fully reproducible as long as the random seed is fixed. In this case, the division between trust and non-trust sets becomes repeatable, and the sample ordering in the non-trust set's data loader is reproducible. Consequently, the new labels assigned to non-trust samples via Equation (2) are perfectly reproducible.

\begin{table}[!t]
\renewcommand{\baselinestretch}{1.0}
\renewcommand\arraystretch{1.0}
\renewcommand\tabcolsep{1pt}
\setlength{\abovecaptionskip}{0pt}
\centering
\caption{ Comparison between Random assignment of labels and using original pseudo-labels for non-trust samples.}
\label{random}
\footnotesize 
\begin{tabular}{@{}l  | cccc@{}}
\toprule
Method                          & Ar$\to$Cl & Ar$\to$Pr & Ar$\to$Rw & Cl$\to$Ar \\
\midrule
Original pseudo-labels          & 80.3  & 91.1  & 90.1  & 83.9  \\
Ours (randomly assigned labels) & 86.7  & 94.2  & 92.8  & 91.5  \\
\bottomrule
\end{tabular}
\end{table}

\subsection{Analysis of Comparison with DATUM}

We compare our method with DATUM~\cite{datum} on the Office-Home dataset. According to the DATUM~\cite{datum} paper, the method consists of three stages: (1) Employing training of DreamBooth~\cite{dreambooth} to personalize the diffusion model by associating a unique token $V_{\ast}$ with the appearance of the target domain. (2) Using the personalized diffusion model to generate a pseudo-target domain. (3) Training an existing UDA framework on the labeled source data and the generated unlabeled pseudo-target data. To align DATUM with the SFDA setting, we modify only the third stage. Specifically, we first train a source model using labeled source data, and then adapt the model to the target domain using the pseudo-target data generated by DATUM. 

The results are shown in Table~\ref{datum}. For reference, we also include the performance of the source model as a baseline for comparison. The results demonstrate that:

(1) Compared with the source model, DATUM achieves better adaptation performance, demonstrating its effectiveness in the SFDA setting. This result suggests that diffusion-based domain adaptation methods like DATUM are capable of significantly improving adaptation performance. We consider DATUM a valuable and insightful work, as its design showcases the potential of leveraging diffusion models  to generate pseudo-target data for SFDA.

(2) Compared with our method, the results demonstrate that DATUM performs significantly worse than our method. This may be due to the following two reasons:
\begin{itemize}
    \item A key factor contributing to the superior performance of our method is the use of our Progressive Refinement Mechanism, which enables the SFDA model to progressively improve its performance through multiple iterations. In contrast, DATUM lacks such a dynamic update mechanism, which may limit its adaptation ability.
    \item We provide the detailed performance trajectory of our method as $r$ increases from 1 to 10 in Table 6. Notably, when $r = 1$, our method still outperforms DATUM. This demonstrates that even without the Progressive Refinement Mechanism, our method remains more effective than DATUM. 
\end{itemize}

(3) The aforementioned results further indicate that the pseudo-target domain generated by our method is better aligned with the real target domain compared to that generated by DATUM.  We attribute this to the following reason. DATUM relies on DreamBooth to learn the appearance of the target domain and map it to a unique token $V_{\ast}$. However, for classification tasks, this process becomes challenging when the true class labels of the target data are unknown. According to the paper of DATUM, during DreamBooth training, the lack of class labels forces the use of vague prompts such as "a photo of a $V_{\ast}$ object". This may cause the learned token to not only capture the domain-specific appearance of the target data, but also absorb semantic information related to the object class and even irrelevant background features. As a result, the generated pseudo-target images may exhibit limited alignment with the true target distribution.

\subsection{Unsupervised Model Selection}

For more complex scenarios, we can also explore other hyperparameter values. In such cases, unsupervised model selection becomes necessary. We employ an unsupervised model selection strategy using the nuclear norm. We extract the softmax output matrix for all target samples and calculate its nuclear norms. Then, we select the model with the largest nuclear norm~\cite{cui2020towards,cui2021fast}. We train models with $E\in[0.01,0.005,0.001]$ and $R$ from 1 to 10 on real-world benchmark VLCS~\cite{vlcs} and apply this model selection metric. The results are shown in Table~\ref{usmvlcs},  comparison results are from~\cite{lee2023few}, where UMS is unsupervised model selection for short. Note that all the methods for comparison report their best target accuracy (according to the labels), while ours performs model selection without knowing target accuracy.
\begin{table}[!t]
\renewcommand{\baselinestretch}{1.0}
\renewcommand\arraystretch{1.0}
\renewcommand\tabcolsep{1pt}
\setlength{\abovecaptionskip}{0pt}
\centering
\caption{Unsupervised Model Selection results on the VLCS dataset.}
\label{usmvlcs}
\footnotesize 
\begin{tabular}{@{}l c | ccccccccccccc@{}}
\toprule
Method             & UMS & C$\to$L  & C$\to$S  & C$\to$V  & L$\to$C  & L$\to$S  & L$\to$V  & S$\to$C  & S$\to$L  & S$\to$V  & V$\to$C  & V$\to$L  & V$\to$S   & Avg  \\
\midrule
No adapt           & \XSolidBrush   & 47.8 & 53.3 & 64.6 & 51.3 & 40.7 & 55.1 & 58.1 & 36.4 & 55.4 & 97.8 & 48.8 & 72.0  & 56.7 \\
SHOT~\cite{shot}       & \XSolidBrush   & 44.6 & 55.7 & 75.9 & 65.3 & 60.8 & 74.8 & 87.7 & 41.1 & 82.7 & 89.8 & 45.8 & 65.7  & 65.6 \\
SHOT++~\cite{shot++}    & \XSolidBrush   & 41.7 & 58.4 & 75.6 & 70.0 & 56.9 & 76.2 & 71.6 & 39.5 & 80.7 & 96.9 & 43.7 & 60.07 & 64.3 \\
AaD~\cite{ada}        & \XSolidBrush   & 37.7 & 57.2 & 75.5 & 59.5 & 48.8 & 67.5 & 84.0 & 34.8 & 72.8 & 43.4 & 40.2 & 54.8  & 56.4 \\
CoWA-JMDS~\cite{cowajmds}  & \XSolidBrush   & 46.1 & 58.4 & 81.1 & 85.6 & 64.1 & 78.2 & 95.9 & 48.1 & 82.7 & 99.4 & 50.8 & 65.3  & 71.3 \\
NRC~\cite{nrc}        & \XSolidBrush   & 39.9 & 55.9 & 75.5 & 64.7 & 54.1 & 74.8 & 78.4 & 40.3 & 82.2 & 90.1 & 41.7 & 62.8  & 63.4 \\
G-SFDA~\cite{gsfda}     & \XSolidBrush   & 42.6 & 54.9 & 73.5 & 82.3 & 51.4 & 72.0 & 74.6 & 45.8 & 82.7 & 88.9 & 49.1 & 64.3  & 65.2 \\
FT~\cite{lee2023few}                 & \XSolidBrush   & 50.1 & 66.7 & 81.1 & 99.7 & 62.2 & 78.0 & 99.8 & 55.5 & 80.5 & 99.7 & 53.0 & 67.0  & 74.5 \\
LP-FT~\cite{lee2023few}              & \XSolidBrush   & 50.5 & 66.7 & 79.5 & 99.7 & 65.6 & 77.6 & 99.7 & 54.1 & 79.5 & 99.7 & 51.2 & 69.8  & 74.5 \\
DPTM (Ours) & \Checkmark   & 88.2 & 96.0 & 90.8 & 97.7 & 97.2 & 99.2 & 99.6 & 88.6 & 89.4 & 99.9 & 92.7 & 91.5  & 94.2 \\
\bottomrule
\end{tabular}
\end{table}

Besides, we also conduct unsupervised model selection experiments on the TerraIncognita dataset~\cite{terrainc}. The experimental setting follows that of the VLCS dataset, and we also use the nuclear norm to select the best model without access to target domain labels. The comparison results are also taken from~\cite{lee2023few}. 

The results are shown in Table~\ref{usmterrainc}. The results demonstrate that, similar to VLCS, our method significantly outperforms all compared methods even when using nuclear norm as the criterion for unsupervised model selection, achieving an average accuracy improvement of 12.4\% over the best-performing comparative method. Notably, in some challenging scenarios such as L100$\to$L43 and L38$\to$L43, where existing methods generally perform poorly, our method achieves substantial gains of 34.8\% and 36.5\%, respectively, even under unsupervised model selection. It is worth noting that all compared methods were evaluated using their best-performing models selected based on target domain accuracy. These results further demonstrate the effectiveness and practicality of our method.

We also acknowledge that there may exist better criteria beyond the nuclear norm. Exploring more effective model selection metrics will be an important direction for our future work, as we believe that a more suitable criterion could further unleash the potential of our method and enhance its practical applicability. 

\subsection{Visualization}

\begin{figure}[!t]
\centering
\includegraphics[width=\textwidth]{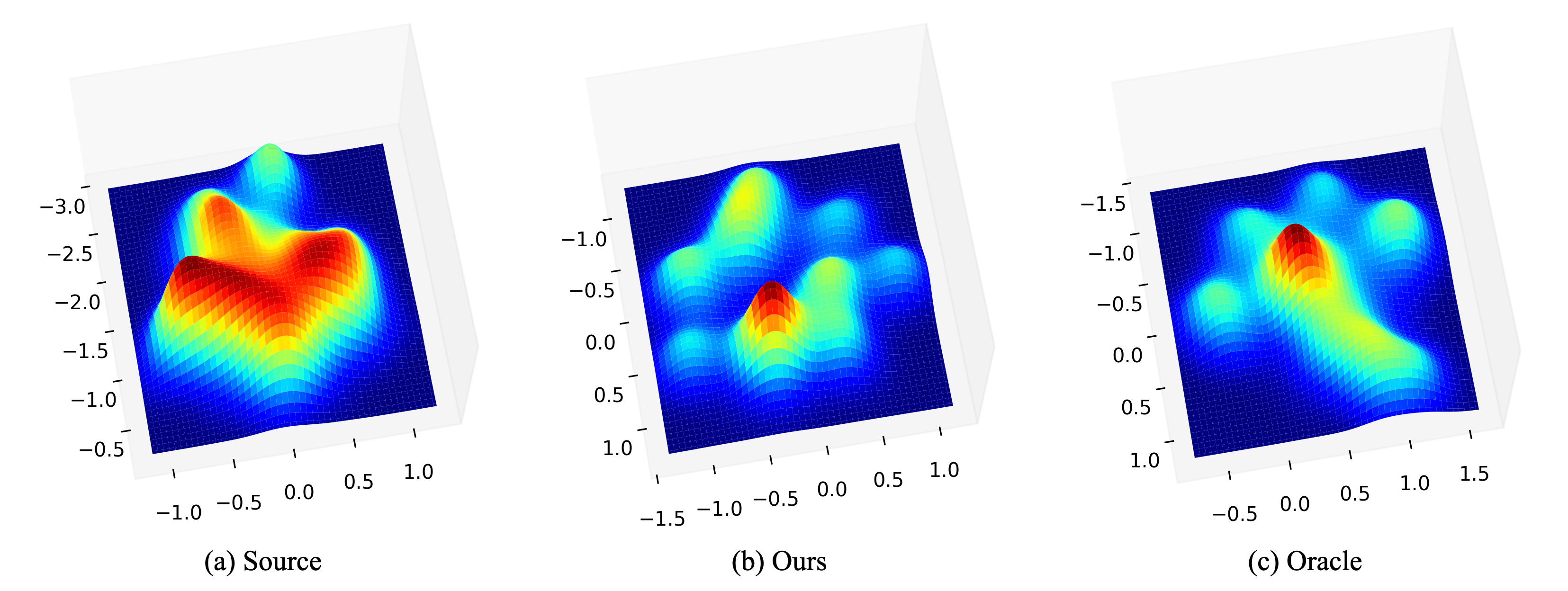} 
\caption{Feature distribution visualization on the Rw$\to$Cl task of the Office-Home dataset.}\label{fig:3dvis} 
\end{figure}

\textbf{Feature Visualization.} We visualize the feature distribution of our DPTM on the Rw$\to$Cl task of the Office-Home dataset using t-SNE, where other SFDA methods perform poorly. The visualization results are shown in Figure~\ref{fig:3dvis}. We compare the results with \textbf{Source} and \textbf{Oracle}, where \textbf{Source} denotes the source model trained only on Rw, and \textbf{Oracle} represents the real target model trained on Cl with ground-truth labels. The visualization is presented as 3D density charts. Following DIFO~\cite{difo}, we use the first 10 categories for clearer viewing. 

\begin{figure}[!t]
\centering
\includegraphics[width=\textwidth]{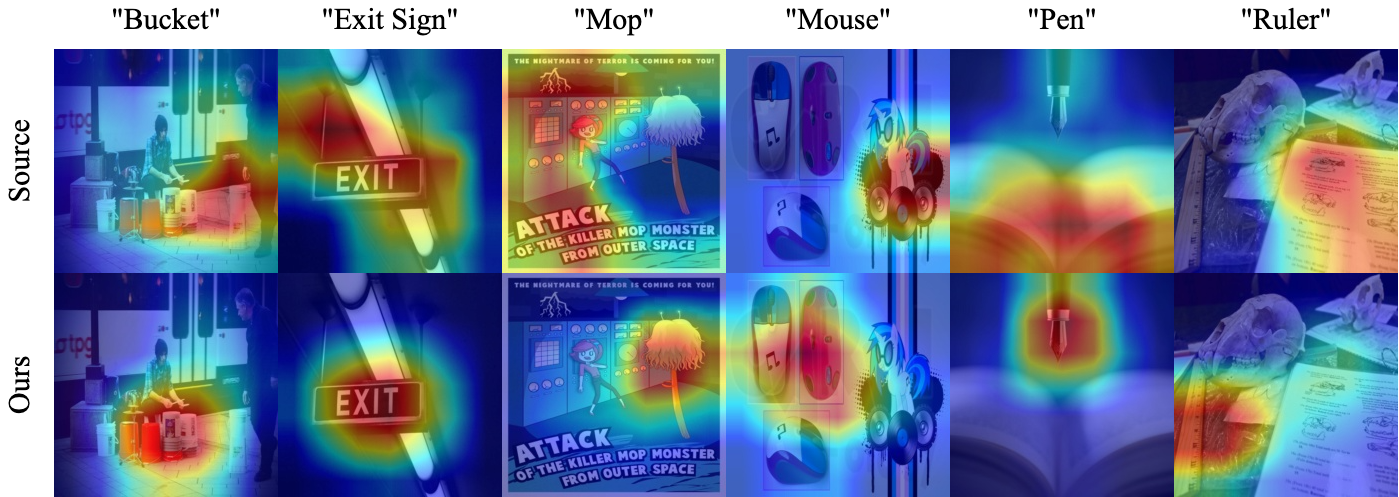} 
\caption{Grad-CAM visualization on the Office-Home dataset.}\label{fig:grad-cam} 
\end{figure}

\textbf{Grad-CAM Visualization.} We further perform Grad-CAM visualization on the Office-Home dataset to analyze the attention behavior of our model. As shown in Figure~\ref{fig:grad-cam}, compared with the source model, our method consistently focuses on the target objects, while the source model tends to attend to irrelevant background regions or even the entire image.

\section{Time and Memory Costs}

\subsection{Theoretical Analysis}
\subsubsection{Time Cost}

The time cost can be mainly divided into two parts:
\begin{itemize}
    \item SFDA model training. For any $r$-th refinement iteration with $r=1,2,...,R$, we train the SFDA model using the trust set and the manipulated non-trust set. The total number of samples of these two sets equals that of the target domain. We use supervised training here. The time cost mainly depends on the scale of the target domain, and remains constant as $r$ grows. We denote it as $t_{SFDA}$.
    \item Sample generation. For any $r$-th iteration refinement with $r=1,2,...,R$, the generation time for each sample remains constant, at about 1 second on a single NVIDIA Tesla V100. The generation time depends on the number of samples of the non-trust set, which is a subset of the target domain. Note that, with the growth of $r$, the size of the non-trust set declines rapidly, as shown in Figure 2. Thus, the total generation time for $R$ iterations equals the total number of generated samples $N_{total}$.
\end{itemize}

  The total training time $t$ can be estimated by $t= R*t_{SFDA}+N_{total}$. We take the 12 tasks in the Office-Home dataset shown in Table 1 as examples. We set $E=0.01$ and $R=10$ and run the tasks on a single NVIDIA Tesla V100. It takes about 7.4 hours for SFDA model training and 3.8 hours for generation to complete the total algorithm on average for each task, demonstrating an acceptable time cost. Note that our model also allows parallel training on multiple GPUs to further reduce the time cost.

\subsubsection{Memory Cost}

The memory needs can be mainly divided into two parts:
\begin{itemize}
    \item SFDA Model Training. During any $r$-th refinement iteration, the memory needs remain constant and approximately equal to those of standard supervised learning on the target domain, and will not accumulate as $r$ grows.
  \item Sample Generation. Our method generates samples sequentially (one sample at a time). Thus, this part only requires sufficient memory to run the Stable Diffusion model.
\end{itemize}

\subsection{Experimental Evaluations}
We compiled comparative experiments on cost analysis, including training time and peak GPU memory usage for our method and other benchmark methods whose official code is publicly available and runnable. We first describe the settings of our cost analysis experiments.

\textbf{Datasets.} We conduct experiments on the Office-Home dataset (moderate scale) and the DomainNet-126 dataset (large scale), as both contain a sufficient number of samples to reasonably evaluate the computational efficiency of different methods.

\textbf{Benchmarked methods.} We successfully ran the following benchmark methods: SHOT, NRC, GKD, AdaCon, CoWA, SCLM, PLUE, TPDS, DIFO, and ProDe. All methods were executed strictly following the instructions provided in their official code repositories.

\textbf{Measurement protocol.} Due to time constraints, we ran each comparative method for one epoch per SFDA task, recorded the training time and peak GPU memory usage during that epoch, and multiplied the time by the number of total epochs provided in their official code repositories to estimate the full training time. As for our own method, we had detailed logs from prior experiments, and we report the actual training time and GPU memory usage based on our full training runs. And we ran all the tasks with a single NVIDIA Tesla V100.

Training time (hours) on the Office-Home dataset is listed in Table~\ref{trainingtimeoffcice}. Office-Home is a medium-scale dataset that contains 4 domains, including Art (Ar, 2427 images), Clipart (Cl, 4365 images), Product (Pr, 4439 images), and Real-World (Rw, 4357 images).

\begin{table}[!t]
\renewcommand{\baselinestretch}{1.0}
\renewcommand\arraystretch{1.0}
\renewcommand\tabcolsep{1pt}
\setlength{\abovecaptionskip}{0pt}
\centering
\caption{Unsupervised Model Selection results on the TerraIncognita dataset.}
\label{usmterrainc}
\scriptsize
\resizebox{\textwidth}{!}{
\begin{tabular}{@{}l c | ccccccccccccc@{}}
\toprule
Method          & UMS & L100→L38 & L100→L43 & L100→L46 & L38→L100 & L38→L43 & L38→L46 & L43→L100 & L43→L38 & L43→L46 & L46→L100 & L46→L38 & L46→L43 & Avg  \\
\midrule
No adapt        & \XSolidBrush   & 26.2     & 20.3     & 27.1     & 29.3     & 31.4    & 31.6    & 24.1     & 44.1    & 38.7    & 33.6     & 21.6    & 22.2    & 29.2 \\
SHOT~\cite{shot}            & \XSolidBrush   & 20.1     & 23.8     & 28.5     & 36.0     & 29.0    & 13.6    & 26.2     & 14.5    & 32.7    & 34.3     & 12.6    & 37.4    & 25.7 \\
SHOT++~\cite{shot++}          & \XSolidBrush   & 29.3     & 22.1     & 25.5     & 22.8     & 31.8    & 18.4    & 33.3     & 22.6    & 25.6    & 35.8     & 13.0    & 44.6    & 27.1 \\
AaD~\cite{ada}             & \XSolidBrush   & 17.2     & 17.4     & 22.1     & 24.6     & 28.1    & 13.3    & 28.9     & 23.3    & 23.1    & 31.6     & 7.4     & 34.6    & 22.6 \\
CoWA-JMDS~\cite{cowajmds}       & \XSolidBrush   & 33.1     & 31.4     & 26.4     & 36.3     & 38.3    & 19.3    & 28.2     & 13.6    & 26.6    & 32.5     & 10.0    & 47.6    & 28.7 \\
NRC~\cite{nrc}             & \XSolidBrush   & 19.3     & 22.7     & 29.5     & 38.5     & 26.9    & 14.9    & 30.8     & 22.6    & 32.2    & 28.9     & 11.0    & 39.0    & 26.4 \\
G-SFDA~\cite{gsfda}          & \XSolidBrush   & 21.6     & 29.1     & 38.2     & 38.4     & 27.0    & 22.4    & 40.9     & 17.4    & 33.3    & 35.0     & 16.3    & 52.6    & 31.0 \\
FT~\cite{lee2023few}              & \XSolidBrush   & 52.4     & 41.7     & 50.0     & 63.8     & 38.6    & 47.8    & 66.2     & 56.7    & 51.4    & 68.9     & 56.7    & 61.2    & 54.6 \\
LP-FT~\cite{lee2023few}           & \XSolidBrush   & 54.3     & 47.5     & 46.9     & 63.6     & 41.3    & 49.0    & 64.2     & 55.9    & 54.4    & 68.4     & 55.7    & 63.8    & 55.4 \\
DPTM (ours) & \Checkmark   & 61.3     & 82.3     & 58.3     & 74.7     & 77.8    & 50.2    & 69.1     & 66.1    & 58.3    & 75.1     & 63.6    & 77.0    & 67.8 \\
\bottomrule
\end{tabular}
}
\end{table}

\begin{table*}[!t]
\renewcommand{\baselinestretch}{1.0}
\renewcommand\arraystretch{1.0}
\setlength{\tabcolsep}{1pt}
\setlength{\abovecaptionskip}{0pt}
\centering
\caption{Training time (hours) on the Office-Home dataset.}
\label{trainingtimeoffcice}
\resizebox{\textwidth}{!}{
\begin{tabular}{@{}l|ccccccccccccc@{}}
\toprule
Method & Ar$\to$Cl & Ar$\to$Pr & Ar$\to$Rw
& Cl$\to$Ar & Cl$\to$Pr & Cl$\to$Rw & Pr$\to$Ar & Pr$\to$Cl & Pr$\to$Rw  
& Rw$\to$Ar & Rw$\to$Cl & Rw$\to$Pr & Avg. \\
\midrule
SHOT            & 3.0            & 2.8            & 8.2            & 1.8            & 3.0            & 8.3            & 1.5            & 2.7            & 7.8            & 1.4            & 2.9            & 3.0            & 3.9           \\
NRC             & 0.9            & 1.4            & 2.4            & 0.9            & 1.2            & 3.9            & 1.3            & 1.7            & 2.6            & 1.0            & 1.2            & 1.4            & 1.7           \\
GKD             & 2.7            & 2.9            & 8.3            & 1.6            & 2.8            & 8.4            & 1.6            & 3.0            & 7.6            & 1.6            & 2.9            & 2.6            & 3.8           \\
AdaCon          & 0.2            & 0.2            & 0.5            & 0.1            & 0.2            & 0.4            & 0.2            & 0.2            & 0.4            & 0.1            & 0.2            & 0.2            & 0.2           \\
CoWA            & 0.9            & 2.4            & 3.5            & 0.6            & 1.2            & 2.6            & 0.8            & 1.6            & 1.5            & 1.0            & 1.2            & 1.5            & 1.6           \\
SCLM            & 2.9            & 3.2            & 7.9            & 1.5            & 3.0            & 9.2            & 1.6            & 2.8            & 8.6            & 1.5            & 2.8            & 2.8            & 4.0           \\
PLUE            & 0.2            & 0.3            & 0.6            & 0.2            & 0.5            & 0.5            & 0.1            & 0.2            & 0.4            & 0.2            & 0.2            & 0.2            & 0.3           \\
TPDS            & 12.3           & 2.7            & 8.0            & 1.6            & 2.9            & 8.4            & 1.5            & 3.0            & 7.8            & 1.4            & 6.0            & 7.5            & 5.3           \\
DIFO            & 6.9            & 7.2            & 22.4           & 4.1            & 7.3            & 25.5           & 4.3            & 7.7            & 26.5           & 4.4            & 7.3            & 7.8            & 11.0          \\
ProDe           & 2.6            & 2.5            & 3.6            & 1.6            & 2.8            & 3.2            & 1.4            & 2.9            & 3.8            & 1.5            & 2.1            & 2.8            & 2.6           \\
\midrule
DPTM (ours) & 11.1           & 9.7            & 16             & 8.7            & 9.9            & 16.2           & 8.6            & 11.3           & 16.2           & 8.6            & 11.1           & 9.7            & 11.4         \\
\bottomrule
\end{tabular}
}
\end{table*}

Similarly, training time (hours) on the DomainNet-126 dataset is listed in Table~\ref{trainingtimedomain}. DomainNet-126 is a large-scale dataset that contains 4 domains, including clipart (C, 18523 images), painting (P, 10212 images), real (R, 69622 images), and sketch (S, 24147 images).

\begin{table*}[!t]
\renewcommand{\baselinestretch}{1.0}
\renewcommand\arraystretch{1.0}
\setlength{\tabcolsep}{2.5pt}
\setlength{\abovecaptionskip}{0pt}
\centering
\caption{Training time (hours) on the DomainNet-126 dataset.}
\label{trainingtimedomain}
\footnotesize
\resizebox{\textwidth}{!}{
\begin{tabular}{@{}l|ccccccccccccc@{}}
\toprule
Method          & C$\to$P  & C$\to$R    & C$\to$S   & P$\to$C  & P$\to$R    & P$\to$S   & R$\to$C  & R$\to$P  & R$\to$S   & S$\to$C  & S$\to$P  & S$\to$R    & Avg.   \\
\midrule
SHOT            & 10.9 & 392.8  & 51.5  & 34.9 & 437.9  & 68.4  & 37.3 & 14.2 & 66.4  & 37.5 & 12.3 & 391.7  & 129.7  \\
NRC             & 1.0  & 5.4    & 2.3   & 3.3  & 9.5    & 4.6   & 2.1  & 1.5  & 3.9   & 2.9  & 1.1  & 4.8    & 3.5    \\
GKD             & 12.2 & 4530.7 & 59.7  & 35.3 & 4134.7 & 67.1  & 37.3 & 10.4 & 72.7  & 38.6 & 13.1 & 4019.8 & 1086.0 \\
AdaCon          & 0.5  & 2.3    & 0.9   & 0.7  & 2.8    & 0.9   & 0.7  & 0.4  & 0.9   & 0.7  & 0.4  & 2.8    & 1.2    \\
CoWA            & 1.4  & 6.2    & 5.3   & 4.8  & 7.8    & 5.4   & 4.4  & 2.8  & 6     & 3.2  & 2.1  & 9.3    & 4.9    \\
PLUE            & 0.5  & 2.8    & 1.0   & 0.8  & 2.8    & 4.6   & 0.8  & 0.5  & 1.0   & 0.8  & 0.5  & 2.8    & 1.6    \\
TPDS            & 11.3 & 383.4  & 52.7  & 39.3 & 426.9  & 68.3  & 39.4 & 14.1 & 69.2  & 40.2 & 13.9 & 406.3  & 130.4  \\
DIFO            & 26.0 & 918.4  & 116.0 & 79.9 & 960.7  & 146.5 & 79.0 & 29.1 & 116.9 & 67.2 & 29.0 & 935.4  & 292.0  \\
ProDe           & 5.1  & 27.1   & 17.0  & 15.8 & 48.1   & 16.7  & 8.8  & 8.1  & 16.9  & 7.8  & 5.2  & 28.5   & 17.1   \\
\midrule
DPTM (Proposed) & 13.7 & 45.3   & 28.5  & 22.2 & 43.6   & 30.1  & 22.2 & 13.7 & 24.8  & 20.4 & 13.3 & 43.5   & 26.8  \\
\bottomrule
\end{tabular}
}
\end{table*}

\begin{table*}[!t]
\renewcommand{\baselinestretch}{1.0}
\renewcommand\arraystretch{1.0}
\setlength{\abovecaptionskip}{0pt}
\centering
\caption{The peak GPU memory usage.}
\label{memory}
\footnotesize
\begin{tabular}{@{}l|cc}
\toprule
Method          & Office-Home & DomainNet-126 \\
\midrule
SHOT            & 7GBytes         & 7GBytes           \\
NRC             & 5GBytes         & 7GBytes           \\
GKD             & 7GBytes         & 8GBytes           \\
AdaCon          & 14GBytes        & 14GBytes          \\
CoWA            & 7GBytes         & 8GBytes           \\
SCLM            & 7GBytes         & -             \\
PLUE            & 13GBytes        & 14GBytes          \\
TPDS            & 7GBytes         & 7GBytes           \\
DIFO            & 7GBytes         & 11GBytes          \\
ProDe           & 12GBytes        & 13GBytes          \\
DPTM (Proposed)      & 10GBytes        & 12GBytes   \\
\bottomrule
\end{tabular}
\end{table*}

As for inference time, following most of the existing benchmark methods, our method uses only the target model (e.g., ResNet-50) for inference on the target domain. No additional modules or auxiliary models are involved during inference, so the inference time of our method is effectively the same as that of other benchmark methods using the same backbone.

Besides, the peak GPU memory usage is listed in Table~\ref{memory}.

\section{Limitation}
Since we employ the pre-trained Stable Diffusion model~\cite{rombach2022high}, the performance of DPTM is inherently constrained by Stable Diffusion's generation capabilities. For the SFDA benchmarks used in our experiments, which primarily consist of common object categories that Stable Diffusion can reliably generate, the framework achieves strong performance. However, when dealing with custom datasets containing categories that are challenging for Stable Diffusion to generate (e.g., specialized medical instruments or rare industrial components), it may be necessary to first collect relevant training data for these specific classes and fine-tune Stable Diffusion accordingly.

\end{document}